\documentclass[10pt,twocolumn,letterpaper]{article}

\usepackage[pagenumbers]{cvpr} % To force page numbers, e.g. for an arXiv version

\usepackage{multirow}
\usepackage{amsmath}
\usepackage{comment}
\usepackage{algorithmic}
\usepackage{algorithm}
\usepackage{bbding} % For \XSolidBrush
\usepackage{bm}
\usepackage{CJKutf8}
\usepackage{array}

\newcommand{\link}[1]{\href{#1}{#1}}

\newcommand{\zhushi}[1]{}
\newcommand{\tb}[1]{\textbf{#1}}
\newcommand{\mb}[1]{\mathbf{#1}}
\newcommand{\Yes}{{\scriptsize{{\Checkmark}}}}
\newcommand{\No}{{-}}

\makeatletter
\newcommand\mythanks[1]{%
  \g@addto@macro\@thanks{\protect\footnotetext[0]{#1}}}
\makeatother

\newcommand{\g}[1]{\textcolor{gray!80}{#1}}

\definecolor{cvprblue}{rgb}{0.21,0.49,0.74}
\usepackage[pagebackref,breaklinks,colorlinks,allcolors=cvprblue]{hyperref}
\def\paperID{397} % *** Enter the Paper ID here
\def\confName{CVPR}
\def\confYear{2025}

\title{Agent-based Video Trimming}

\author{
  Lingfeng Yang$^{1\dagger}$, Zhenyuan Chen$^{2\dagger}$, Xiang Li$^{3,2 *}$, Peiyang Jia$^{4}$, Liangqu Long$^{4}$, Jian Yang$^{1 *}$\mythanks{$^{*}$Corresponding authors, $^{\dagger}$Equal contributions.}
  \\
  {\normalsize
  $^{1}$Nanjing University of Science and Technology,
  $^{2}$VCIP, CS, Nankai University, 
  $^{3}$NKIARI, Shenzhen Futian,   
  $^{4}$Insta360}
  \\
  {\tt\small \{yanglfnjust, csjyang\}@njust.edu.cn, \{zhenyuanchen, xiang.li.implus@\}@nankai.edu.cn} \\
  {\tt\small jiapeiyang@insta360.com, liangqu.long@gmail.com} \\
}

\begin{document}
\maketitle

\begin{abstract}
As information becomes more accessible, user-generated videos are increasing in length, placing a burden on viewers to sift through vast content for valuable insights. This trend underscores the need for an algorithm to extract key video information efficiently. Despite significant advancements in highlight detection, moment retrieval, and video summarization, current approaches primarily focus on selecting specific time intervals, often overlooking the relevance between segments and the potential for segment arranging. In this paper, we introduce a novel task called \textbf{V}ideo \textbf{T}rimming (\textbf{VT}), which focuses on detecting wasted footage, selecting valuable segments, and composing them into a final video with a coherent story. To address this task, we propose \textbf{A}gent-based \textbf{V}ideo \textbf{T}rimming (\textbf{AVT}), structured into three phases: Video Structuring, Clip Filtering, and Story Composition. Specifically, we employ a Video Captioning Agent to convert video slices into structured textual descriptions, a Filtering Module to dynamically discard low-quality footage based on the structured information of each clip, and a Video Arrangement Agent to select and compile valid clips into a coherent final narrative. For evaluation, we develop a Video Evaluation Agent to assess trimmed videos, conducting assessments in parallel with human evaluations. Additionally, we curate a new benchmark dataset for video trimming using raw user videos from the internet. As a result, AVT received more favorable evaluations in user studies and demonstrated superior mAP and precision on the YouTube Highlights, TVSum, and our own dataset for the highlight detection task. The code and models are available at \link{https://ylingfeng.github.io/AVT}.
\end{abstract}

\begin{figure}[t]
    \centering
    \includegraphics[width=\linewidth]{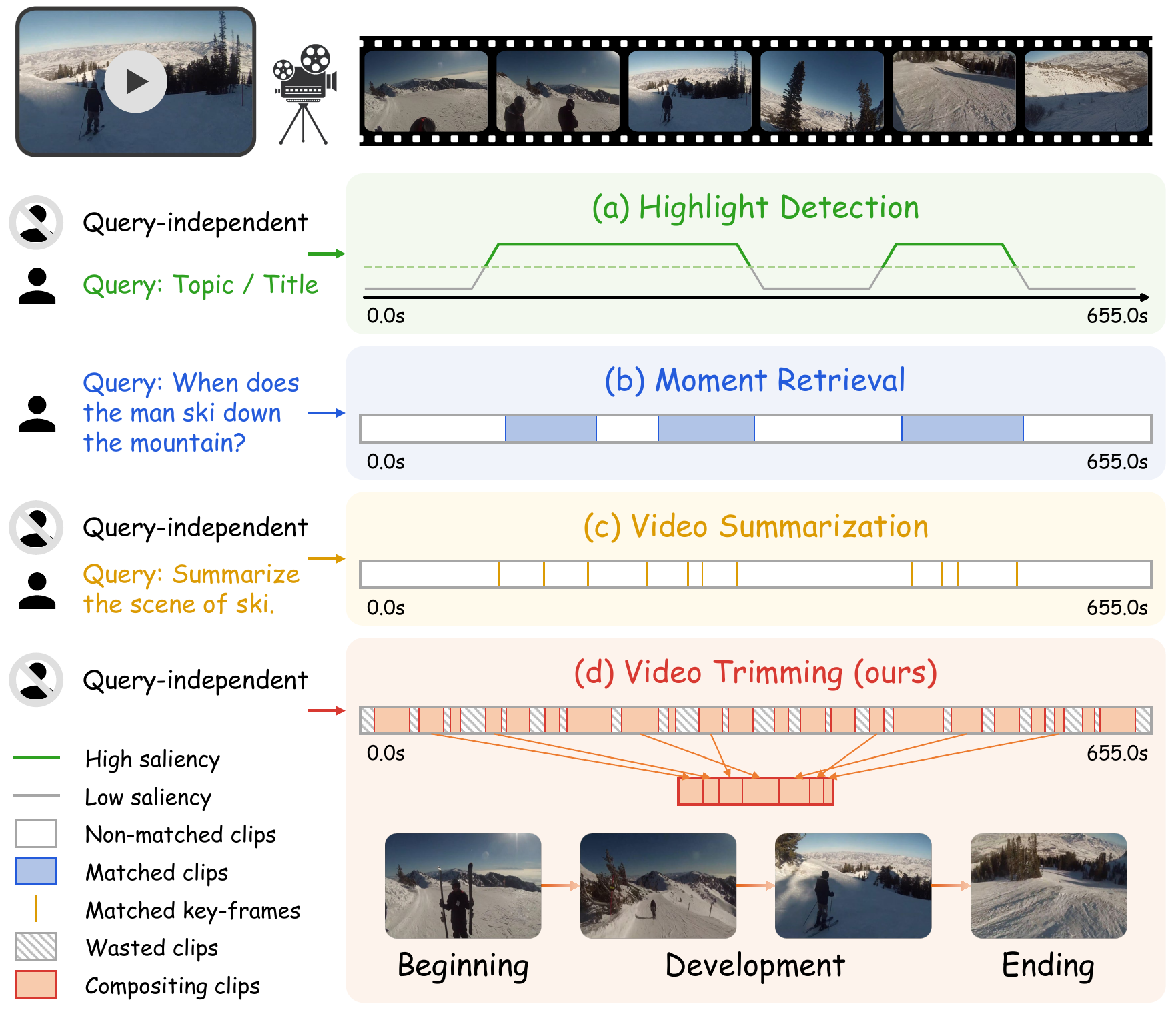}
    \caption{
    A comparison between our new task and existing video tasks: (a) Highlight Detection retrieves clips above a saliency threshold. (b) Moment Retrieval identifies the start and end for intervals related to a given query. (c) Video Summarization extracts keyframes for each theme of the video. 
    (d) Video Trimming addresses more than just a retrieval task by also filtering wasted footage and logically composing the selected segments.
    }
    \label{fig_tasks_overview}
\end{figure}

\begin{figure*}[t]
    \centering
    \includegraphics[width=\textwidth]{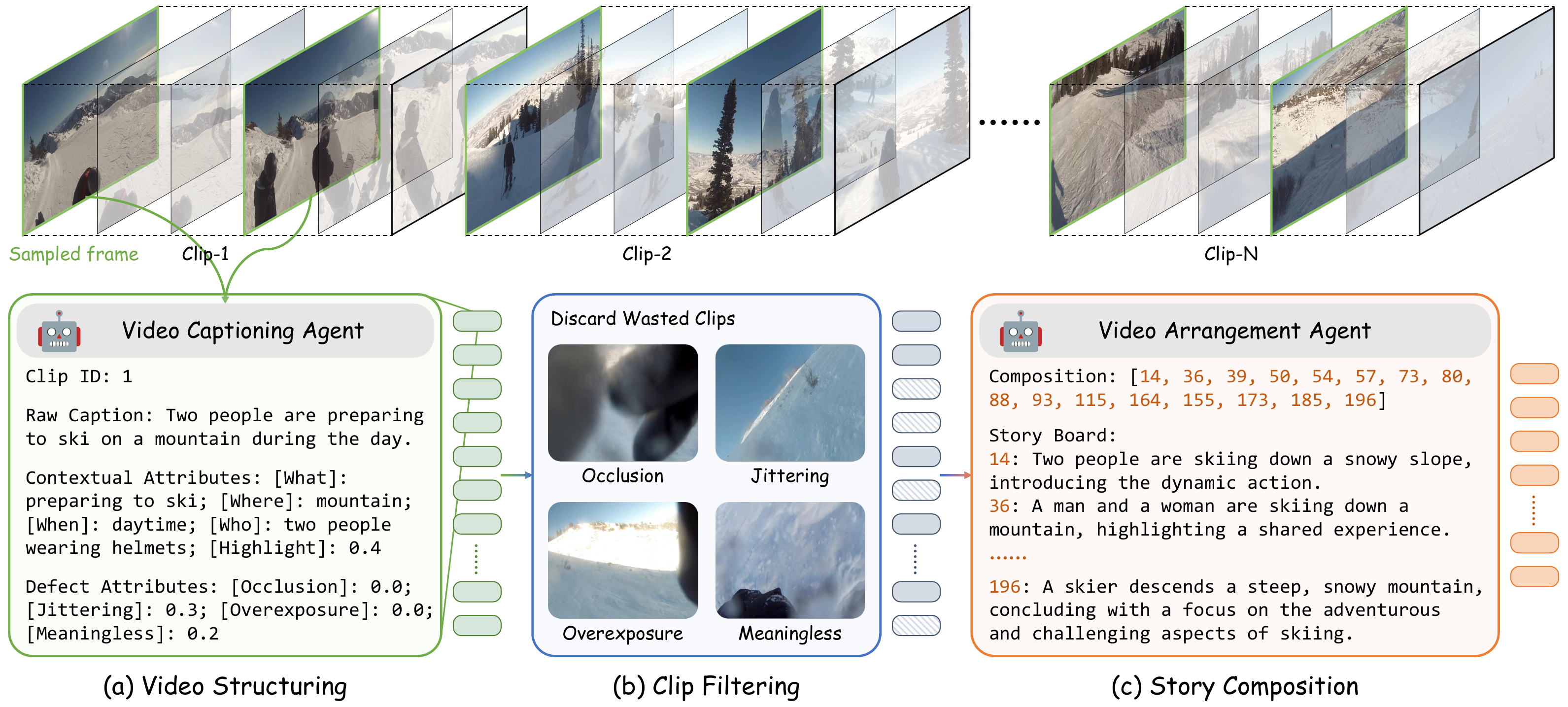}
    \caption{
    The overall framework of AVT. The approach first (a) converts sampled video content into structured captions and attributes, then (b) discards defective clips, and finally (c) organizes the remaining clips into a coherent final cut.
    }
    \label{fig_main_framework}
\end{figure*}

\section{Introduction}\label{sec:intro}
With the ever-increasing volume of visual information and daily video content, there is an urgent need for algorithms that can perform video comprehension and effectively distill critical information from redundant content.

Understanding the video is essential to reducing excessive content. Substantial advancements are made in the field of video understanding~\cite{zhao2023lavila, yang2023vid2seq, maaz2023videochatgpt, damonlpsg2023videollama, lin2023videollava, suris2023vipergpt}. Building on these foundations, methods for highlight detection~\cite{sun2014ranking, song2015tvsum, lei2021qvhighlights, xiong2019less, moon2023query} focus on predicting saliency scores to identify and extract significant segments from videos, thereby reducing the amount of redundant information. Moment retrieval~\cite{krishna2017dense, anne2017localizing, gabeur2020multi, mithun2019weakly, moon2023query} seeks to identify specific moments in a video that correspond to a given query, while video summarization~\cite{apostolidis2021combining, zhou2018deep, gygli2014creating} compiles keyframes to capture detected themes within the video.

However, current approaches focus solely on content extraction and retrieval, without considering the relationships and coherence between segments. To overcome this limitation, we propose a novel task for the first time: Video Trimming (VT). This task involves not only selecting high-saliency segments but also filtering out wasted footage and arranging the remaining clips, resulting in a logically structured and cohesive video output. A comparison with existing tasks is shown in Fig.~\ref{fig_tasks_overview}.

To establish a feasible baseline for this task, we consider leveraging the concept of agents. Recent observations indicate that multimodal large language models~\cite{gpt4, gemini, internvl, claude3} (MLLMs) exhibit robust capabilities in in-context communication and formatting interactions, positioning them as effective, training-free agents for video understanding tasks such as video captioning~\cite{li2022mplug, yang2023vid2seq} and question answering~\cite{suris2023vipergpt, yu2024self, wang2022internvideo}.
Agent-based approaches utilize MLLMs with specially designed prompting pipelines to summarize videos into text and organize the summaries based on given inputs~\cite{wang2024videoagent, wang2023chatvideo, lin2023mmvid}. Alternatively, these models can also function as controllers, coordinating various executors, such as tracking and captioning systems, to address complex multimodal video tasks~\cite{suris2023vipergpt, yang2023mm, wu2023visual, liu2023interngpt, gao2023assistgpt}.

Despite the extensive use of agent-based approaches in video understanding tasks, this study aims to harness their capabilities to develop the first video trimming algorithm. By integrating MLLMs, our approach targets the editing of long-form videos, which may extend up to half an hour or longer, and trims them into shorter, viewable final cuts. Specifically, Agent-based Video Trimming (AVT) presents innovations across three key phases: Video Structuring, Clip Filtering, and Story Composition (Fig.~\ref{fig_main_framework}). In the Video Structuring phase, the video is divided into smaller units. A Video Captioning Agent then converts these units into structured textual descriptions, enabling detailed semantic analysis of each segment. Therefore, the subsequent processes are free of visual content, allowing for faster and more efficient results. In addition to generating basic descriptions, we incorporate attributes that denote defects in video segments, such as occlusion, jittering, overexposure, and meaningless content, to evaluate frame quality. The Clip Filtering phase employs a dynamic module to select useful clips by analyzing structured textual descriptions, differentiating between valuable and irrelevant content. In the Story Composition phase, a Video Arrangement Agent assembles the selected clips into a cohesive final video, ensuring a coherent and engaging narrative flow.

Additionally, we design a Video Evaluation Agent to assess the quality of the trimmed videos. We also create a new benchmark for video trimming, which includes raw user videos annotated with waste and highlight labels. For evaluation, we conduct user studies and quantitative experiments on zero-shot highlight detection across three benchmarks: YouTube Highlights~\cite{sun2014ranking}, TVSum~\cite{song2015tvsum}, and our dataset.

Our contributions can be summarized as follows:
\begin{itemize}
\item[$\bullet$] To the best of our knowledge, we are the first to introduce Video Trimming (VT), a novel task that extracts key intentions from long-form user-shot videos to generate condensed videos with a coherent storyline.
\item[$\bullet$] To establish a baseline, we propose the AVT algorithm, which converts video content into structured descriptions, filters out wasted clips, and arranges the selected clips into a coherent final narrative video.
\item[$\bullet$] We propose a new video trimming benchmark by integrating web videos and using a Video Evaluation Agent to assess video quality alongside human evaluation.
\item[$\bullet$] Our method demonstrates superior performance in video trimming and zero-shot highlight detection, as evidenced by both user studies and various benchmarks.
\end{itemize}

%-------------------------------------------------------------------------
%-------------------------------------------------------------------------
%-------------------------------------------------------------------------
%-------------------------------------------------------------------------
%-------------------------------------------------------------------------

\section{Related Work}\label{sec:related_work}
\subsection{Video Understanding}
Leveraging MLLMs, a wide range of research studies are conducted to advance video understanding. The existing methods encompass one or multiple tasks, such as video question answering~\cite{wang2022internvideo,zhao2023lavila,luo2023valley,li2023videochat,yang2023vid2seq}, long video understanding~\cite{weng2024longvlm,weng2024longvlm}, and moment localization~\cite{yu2024self}. InternVid~\cite{wang2023internvid} builds a large-scale video-text model through contrastive learning and multi-stage pre-training. InternVideo~\cite{wang2022internvideo,wang2024internvideo2} uses multimodal data to scale video understanding. Models like LaViLa~\cite{zhao2023lavila} and Valley~\cite{luo2023valley} improve video-based instruction understanding via fine-tuning. Merlin~\cite{yu2024merlin} and MovieChat~\cite{song2024moviechat} enhance video question answering and long video comprehension. PaLM-E~\cite{driess2023palm} integrates real-world sensory data into language models, while SeViLA~\cite{yu2024self} uses keyframe localization for event prediction. Vid2Seq~\cite{yang2023vid2seq} and VideoChat~\cite{li2023videochat} achieve chat-centric video understanding through fine-tuning. Recently, models like LongVLM~\cite{weng2024longvlm} and VTimeLLM~\cite{huang2024vtimellm} improve comprehension of long videos by segmenting and identifying moments.  
In contrast to addressing generic video tasks, our method specifically targets video trimming, utilizing foundation models as integral components.

\subsection{Video Agent}
Existing video agent methods fall into two main development types. 
The first type leverages large language models (LLMs) in combination with external tools and code executors. DoraemonGPT~\cite{yang2024doraemongpt} enhances VQA tasks by using symbolic memory for better retrieval and summarization. Similarly, InternGPT~\cite{liu2023interngpt} improves reasoning through interactive polling, while MM-ReAct~\cite{yang2023mm} extends the REACT~\cite{yao2023react} mechanism to multimodal tasks. Video ChatCaptioner~\cite{chen2023video} strengthens video understanding with multi-agent iterative polling.
The second targets specific video understanding tasks by converting video content into a textual corpus for subsequent analysis. AssistGPT~\cite{gao2023assistgpt} improves VQA and moment retrieval through a cycle of planning, execution, inspection, and learning. ChatVideo~\cite{wang2023chatvideo} structures video content into a text database for efficient querying, while LLoVi~\cite{zhang2023llovi} focuses on fine-grained VQA and interval retrieval using captions. MM-Vid~\cite{lin2023mmvid} treats multimodal information as textual data, and VideoAgent~\cite{wang2024videoagent} improves moment retrieval with iterative polling and similarity matching.
The first type of video agent is inadequate for video trimming, as they lack specialized models or tools to solve this task. For the second type, although video retrieval agents can extract segments based on queries, they often neglect crucial global content, compromising video coherence. In contrast, our approach is the first to tackle this challenge by creating an innovative video processing pipeline that incorporates a video agent system.

\subsection{Video Temporal Grounding}
This task aims to ground target clips from videos in the form of continuous intervals or discrete keyframes, encompassing highlight detection, moment retrieval, and video summarization.  
Firstly, \emph{Highlight Detection}~\cite{sun2014ranking, song2015tvsum, gygli2016video2gif, xiong2019less, badamdorj2021joint, badamdorj2022contrastive} (HD) predicts saliency scores to extract highlight segments, capturing key visual or contextual moments. However, these highlight-based methods lack the temporal context and event relationships needed for coherent video trimming.  
Next, \emph{Moment Retrieval}~\cite{hong2020mini, anne2017localizing, krishna2017dense, gao2017tall, mithun2019weakly} (MR) selects moments based on a given query. Datasets such as DiDeMo~\cite{anne2017localizing}, ActivityNet Caption~\cite{krishna2017dense}, and Charades-STA~\cite{gao2017tall} provide regional captions for video slices to facilitate retrieval tasks.  
Further, methods such as Moment-DETR~\cite{lei2021qvhighlights}, QD-DETR~\cite{moon2023query}, TR-DETR~\cite{sun2024trdetr}, UniVTG~\cite{lin2023univtg}, and UVCOM~\cite{xiao2024bridging} aim to address both moment and saliency prediction through reciprocal module designs. However, the retrieved segments often lack comprehensive video coverage and context, and they require a prior user query.
Finally, \emph{Video Summarization} condenses videos by selecting key shots that best represent the content of the raw video. Generic summarization~\cite{gygli2014creating, song2015tvsum, mahasseni2017unsupervised, jiang2022joint} relies solely on visual cues, while query-based summarization~\cite{sharghi2017query, nalla2020watch, wu2022intentvizor, lin2023univtg} allows users to tailor the summary by specifying text-based keywords. Although summarization condenses the video, the selected segments are discrete and fail to create a coherent, viewable video.  
Above all, the video temporal grounding task focuses solely on segment selection, often neglecting narrative flow. In contrast, our proposed video trimming task emphasizes both segment selection and composition, preserving narrative integrity while shortening video duration.

%-------------------------------------------------------------------------
%-------------------------------------------------------------------------
%-------------------------------------------------------------------------
%-------------------------------------------------------------------------
%-------------------------------------------------------------------------

\section{Method}\label{sec:method}
In this section, we introduce the Video Trimming (VT) task, which extends beyond highlight selection to include filtering out extraneous footage and creating cohesive, narratively coherent video outputs. To establish a baseline for this task, we propose Agent-based Video Trimming (AVT), an algorithm that leverages MLLMs~\cite{gpt4, gemini, internvl, claude3} as training-free agents across three phases: Video Structuring, Clip Filtering, and Story Composition. Finally, we present an agent-based evaluation metric to assess the final cut, complemented by a user study.

\begin{figure}[t]
    \centering
    \includegraphics[width=\linewidth]{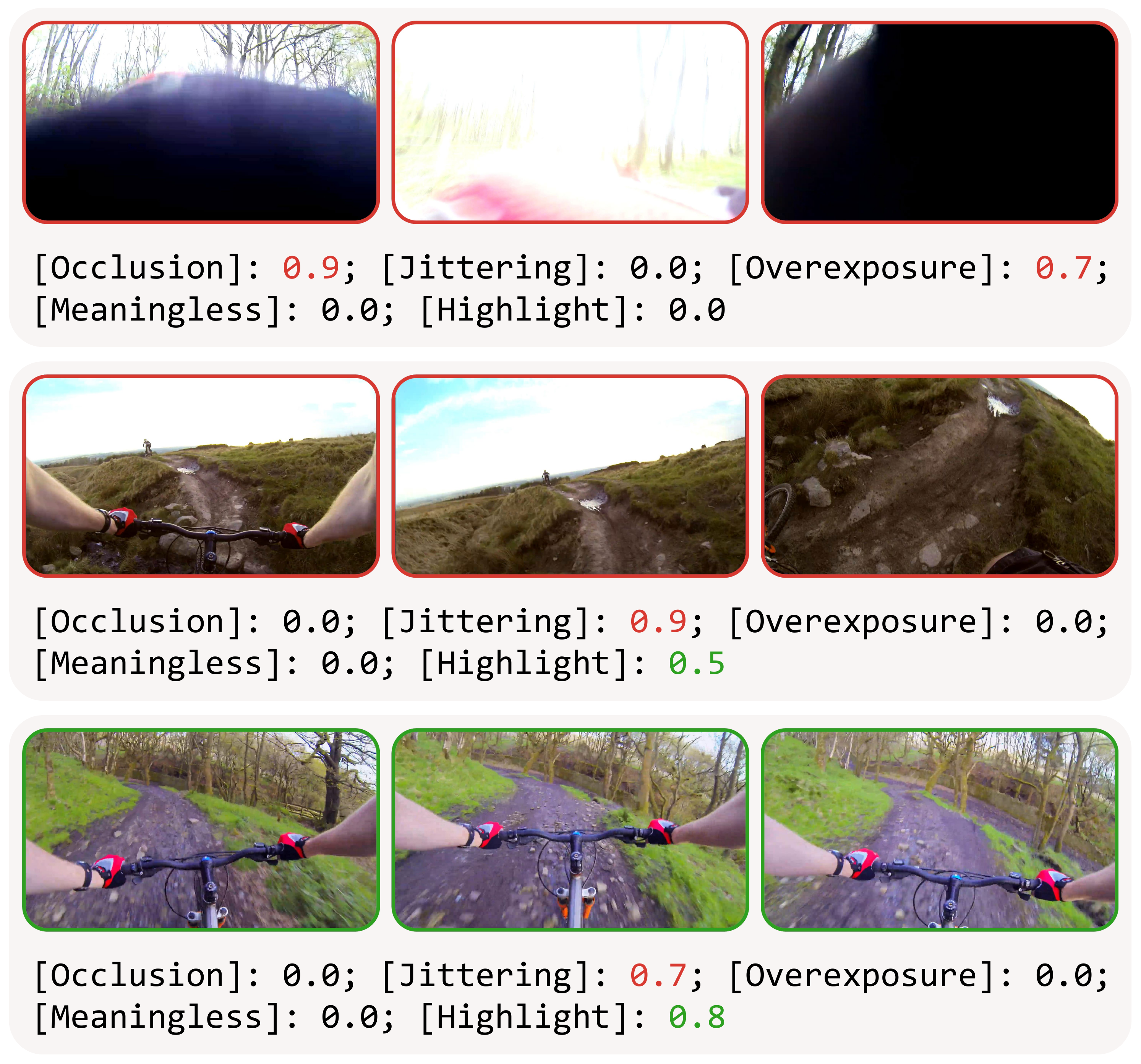}
    \caption{
    Keyframes from a mountain biking video. Clips marked with red boxes are discarded due to higher defect scores, while clips with green boxes are selected despite minor shaking, as they highlight the dynamic scene of cycling on a mountain path.
    }
    \label{fig_dynamic_filter}
\end{figure}

\subsection{Video Structuring}
Recent multimodal video tasks perform video understanding by extracting information from visual contexts to derive semantic features~\cite{li2023videochat, chen2023videollm, damonlpsg2023videollama, lin2023videollava, maaz2023videochatgpt} or directly generating descriptive text~\cite{lin2023mmvid, wang2024videoagent, wang2023chatvideo}. In our case, we adopt the latter to ensure compatibility with multimodal agents such as GPT-4~\cite{gpt4} or Gemini~\cite{gemini}. Notably, once videos are processed as text, subsequent operations become independent of the visual content, which enhances processing speed and reduces computational costs by handling only text inputs.

To structure the video content, we divide the frames into clips, each with a default duration of 3 seconds. For each clip, frames are sampled at a rate of one per second to provide visual inputs. In contrast to previous works that merely derive generic descriptions of video content, we aim to assess the quality of each clip based on its filming characteristics. Empirically, a handheld recording may contain flawed footage, such as occlusion of the target by obstacles or excessive camera shake, both of which detract from the viewing experience. These defects are only weakly related to the events depicted in the visual cues, therefore they are typically not captured by general summaries. As a result, we need to specifically extract defect attributes to enable a thorough assessment of clip quality. To this end, we identify four primary defects: occlusion, jittering, overexposure, and meaningless content, as illustrated in Fig.~\ref{fig_main_framework}~{(b)}. Specifically, meaningless content refers to scenes characterized by simple transitions, empty shots devoid of substantive information, or extended static frames.

Moreover, to handle lengthy raw videos, it is essential to eliminate redundant segments while preserving highlights and engaging parts. Although raw captions offer detailed information, they are inadequate for segment trimming, as similar visual content can yield diverse textual descriptions. To address this, we introduce contextual attributes that summarize video content across four dimensions: what, where, when, and who, offering brief insights into the activity, location, timing, and potential characters. Additionally, we design a ``Highlight'' attribute to measure the overall excitement level of each clip.

To obtain the aforementioned attributes, we employ the MLLMs as the Video Captioning Agent to extract the Raw Caption, Contextual Attributes, and Defect Attributes for every clip, as shown in Fig.~\ref{fig_main_framework}~{(a)}. The $Clip \ ID$ follows a natural sequence related to the length of the video. We expect the structured text information to consist of short sentences or phrases, except for the ``Highlight'' attribute and all defect attributes, which should return a float value ranging from 0 to 1, indicating the degree to which a clip is a highlight or exhibits specific flaws. If the value is 0, it indicates a negative attribute. These scores will be utilized in the next section to dynamically filter out wasted clips.

\begin{algorithm}[t]
    \caption{Dynamic Filtering Algorithm}
    \label{alg_string_processing}
    \renewcommand{\algorithmicrequire}{\textbf{Input:}}
    \renewcommand{\algorithmicensure}{\textbf{Output:}}
    \begin{algorithmic}[1]
        \REQUIRE List of Keys $keys$, List of Numbers $nums$
        \ENSURE $(\text{filter\_flag}, \text{highlight\_flag}, \text{highlight\_score})$
        
        \STATE Initialize $score \gets 0$, $max\_key \gets \text{None}$
        
        \FOR{each $(key, num)$ \textbf{in} \textbf{zip}($keys$, $nums$)}
            \IF{$num \geq score$}
                \STATE $score \gets num$
                \STATE $max\_key \gets key$
            \ENDIF
        \ENDFOR
        
        \IF{$max\_key = \text{[Highlight]}$} 
            \RETURN $(\text{False}, \text{True}, score)$
        \ELSE 
            \RETURN $(score \neq 0, \text{False}, score)$
        \ENDIF
    \end{algorithmic}
\end{algorithm}

\begin{figure*}[t]
    \centering
    \includegraphics[width=\textwidth]{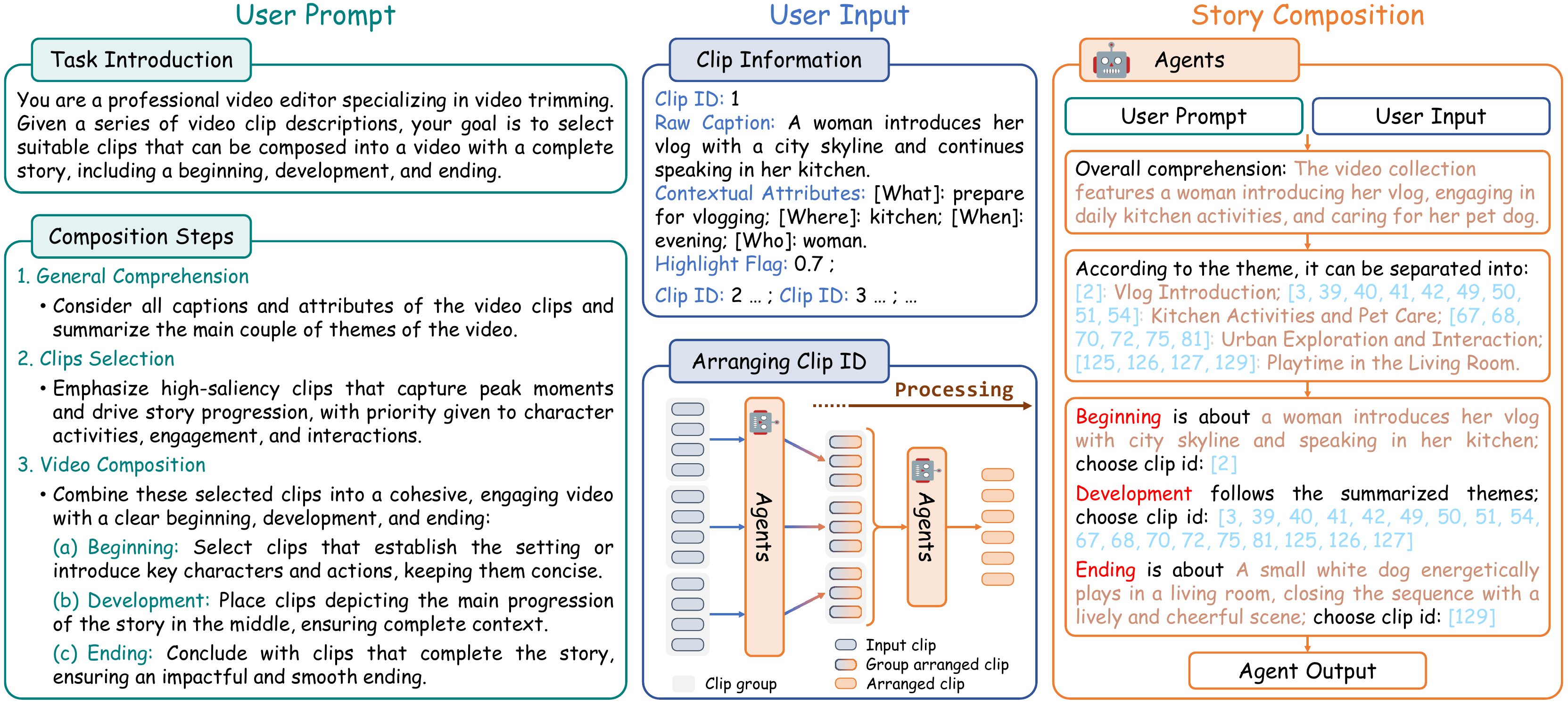}
    \caption{
    The overall framework of the story composition phase begins with inputting the task introduction to the LLM to generate a CoT of the composition steps. With minor adjustments, we call the Video Arrangement Agent, prompted with the refined user input, to sequentially select clips and arrange the story. The output consists of the selected clip indices and segmented story content.
    }
    \label{fig_story_composition}
\end{figure*}

\subsection{Clip Filtering}
In contrast to existing moment retrieval methods that score videos based on their alignment with a specific query, we focus exclusively on visual properties to assess the quality of the footage.
We gather defect attributes and the highlight score from the output of the Video Captioning Agent, formatted as strings: ``\emph{[Occlusion]: 0.8; [Jittering]: 0.7; [Overexposure]: 0.0; [Meaningless]: 0.0; [Highlight]: 0.9}''. This output includes four defect indicators and one highlight score, used as inputs for the filtering mechanism.

Specifically, a common practice is to filter out all clips with any negative defect scores greater than zero. However, this is not always practical. In videos filmed from a first-person perspective, camera shake is inevitable, and the agent would mark clips as jittery, resulting in the exclusion of useful content. To address this, we introduce a positive indicator to balance against the negative ones. The hypothesis is to ensure the algorithm focuses more on the intense video content itself rather than on minor filming imperfections. Based on this strategy, a clip is selected as valid for the final composition only when its ``Highlight'' score exceeds all defect scores. This mechanism termed the Dynamic Filter, balances content richness with filming defects. As shown in Fig.~\ref{fig_dynamic_filter}, we visualize diverse frames from clips to demonstrate how this filtering rule is applied.

For a detailed understanding, we present the string processing algorithm in {Alg.}~\ref{alg_string_processing}. The algorithm processes the structured data by parsing it into attribute-score pairs. For the returned value, the defect flag determines whether a clip should be filtered out, while the highlight flag and score are further used in the next phase of story composition.

%-------------------------------------------------------------------------

\subsection{Story Composition}
In this section, we introduce an agent for story composition, which arranges filtered clips into a coherent order. For the user prompt, we present the video agent with an introduction to the task and involve Chain of Thought~\cite{wei2022chain} (CoT) to generate the video composition steps, taking into account the global concept, clip selection, and composition arrangement (Fig.~\ref{fig_story_composition}). We denote the entire user prompt as $P$. Then, assuming we obtain $M$ valid clips with indices $C = \{C_1, C_2, \ldots, C_M\}$, we format the user input $I$ by concatenating the structured information as follows:
\begin{equation}
\begin{aligned}
    I = \ & \{\{Clip \ ID\}_k, \{Highlight \ Flag\}_k \ (\{Score\}_k), \\
    &\{Contextual \ Attributes\}_k,  \\
    &\{Raw \ Caption\}_k\} \ | \ _{Clip-k \ \{C_1\sim C_M\}},
\end{aligned}
\end{equation}
where the $Highlight \ Flag$ and the corresponding $Score$ are derived from the filtering phase, while the remaining information is obtained from the structuring phase. Next, we prompt the Video Arrangement Agent with $P$ and $I$, expecting the output storyline to consist of the preferred sequence, which is generated through a mapping operation from each sentence to its corresponding clip index, arranged in a sensible order.
After processing, we expect outputs that include the sequence of composite clips, denoted as $C^t$, along with the narrative and the reasoning behind their organization, as illustrated in Fig.~\ref{fig_main_framework}~{(c)}. The composition phase may be iterated until the desired length of the output video is achieved. It is crucial to note that processing too many clips at once can result in ambiguous outputs, as LLMs face difficulties with long contexts and are prone to distraction~\cite{shi2023large}. Therefore, we group the clips and call the agent in parallel for the initial processing. As the number of clips decreases, all information is integrated into the final composition.

Subsequently, we map the final clip indices back to their respective video durations and assemble them to form the final video. It is important to note that not all clips will be selected, and the sequence of clips may not strictly adhere to chronological order; rather, they will be arranged according to the storyline organized by the agent. For details on the prompt design, see the \textit{Supplementary Materials}.

Existing approaches primarily focus on image-text matching, emphasizing the accuracy of retrieved video moments and aiming to maximize their overall salience. However, in the context of video trimming, this is not the sole objective. A well-told story should prioritize the most prominent activities, while also incorporating the introductory and concluding segments. Although the beginning and end may appear less prominent than the main content, they are essential. We emphasize the importance of frame composition, which is achieved not only through the arrangement of clips but also by incorporating slightly less highlighted segments. These segments can effectively serve as transitions, bridging the preceding and subsequent content.

\subsection{Final Cut Evaluation}\label{sec_video_evaluation_agent}
As indicated in G-Eval~\cite{liu2023g}, LLM-based evaluators possess the ability to assess the quality of natural language generation. We extend this automatic evaluation to multimodal tasks by utilizing LLM as the Video Evaluation Agent to assess final videos.
Directly prompting MLLMs for aesthetic evaluation often aligns poorly with human assessments~\cite{fu2023gptscore,wang2023chatgpt}. To improve precision, we define evaluation criteria and create CoT instructions for the Video Evaluation Agent. The criteria, rated from 1 to 5, include Material Richness, assessing diversity and narrative coherence; Appeal, measuring engagement, length, and entertainment; Exciting Segments, evaluating highlight quality and frequency; and Amount of Wasted Footage, considering irrelevant content, with higher scores indicating fewer distractions and better viewing. The Video Evaluation Agent uses only video content as input and outputs scores for each metric, along with justifications. For example: ``\emph{[Material Richness]: \{Reason\} (2.5); [Appeal]: \{Reason\} (3.0); [Exciting Segments]: \{Reason\} (3.5); [Amount of Wasted Footage]: \{Reason\} (2.0);}.'' We calculate the average of all scores to determine the final rating of a video cut.

%-------------------------------------------------------------------------
%-------------------------------------------------------------------------
%-------------------------------------------------------------------------
%-------------------------------------------------------------------------
%-------------------------------------------------------------------------

\begin{table}[t]
\centering
\renewcommand\tabcolsep{3pt}
    \centering
    \resizebox{\linewidth}{!}{%
    \begin{tabular}{@{}lccccc|c@{}}
        \toprule
        \tb{Method} & Richness & Appeal & Excitement & Wasted & \tb{Overall} & \tb{Agent} \\
        \midrule
        {UniVTG}~\cite{lin2023univtg}    & $6.41$ & $7.15$ & $4.74$ & $6.04$ & $6.30$ & $3.03$\\
        {UVCOM}~\cite{xiao2024bridging}  & $6.15$ & $7.12$ & $4.69$ & $6.47$ & $6.23$ & $2.91$ \\
        AVT (ours)                       & $\mb{7.21}$ & $\mb{7.78}$ & $\mb{5.57}$ & $\mb{6.72}$ & $\mb{7.15}$ & $\mb{3.32}$\\
        \bottomrule
    \end{tabular}
    }%
    \caption{
    User study through blind testing, using a scale from 1 to 10, of different methods on the video trimming dataset, comprising 30 final cuts from 42 raw videos and involving 17 participants.
    }
    \label{tab_blind_test_avt}
\end{table}%

\section{Experiments}
In this section, we first introduce the dataset and implementation details. We then compare the quality of the trimmed videos, along with quantitative experiments on highlight detection. Next, ablation studies on the main components of AVT are presented. Lastly, we provide visualizations of the results and case studies for further discussion.

\subsection{Datasets}
\noindent\textbf{Existing Dataset.} The YouTube Highlights~\cite{sun2014ranking} and TVSum~\cite{song2015tvsum} datasets are two well-established benchmarks for evaluating video temporal grounding tasks. We tested on 20\% of the data, following the same split as in~\cite{lin2023univtg, liu2022umt}. For evaluation metrics, we followed the methodology described in~\cite{liu2022umt}, using mAP for the YouTube Highlights dataset and Top-5 mAP for the TVSum dataset.

\noindent\textbf{Video Trimming Dataset.} 
Additionally, we collect web-crawled videos from YouTube and construct a benchmark specifically for video trimming. We categorize common user video types into three groups: daily life, sports, and travel vlogs. For each category, we select 10 video uploaders and choose one or more videos filmed around a consistent event, meaning the algorithm may be requested to composite video cuts from multiple source videos. In total, we compile 30 topics with 42 videos, each averaging 10 minutes in length. We annotate each video with four ranks of scores: 0 for wasted, 1 for ambiguous, 2 for normal, and 3 for highlight footage. Detailed illustrations are provided in the \textit{Supplementary Materials}.

\subsection{Implementation Details}
To enhance multimodal interaction capabilities and ensure restricted output formatting, we implement all agents in our AVT algorithm using the GPT-4o model~\cite{gpt4o}. For the visual inputs, the video is divided into 3-second segments, with one keyframe sampled per second by default. All frame images are resized to 512 pixels on the shorter side. The text inputs include both our designed prompting instructions and the structured video captions and attributes. This configuration results in approximately 153,000 input image tokens, 100,000 input text tokens, and 20,000 output text tokens for a single 10-minute raw video, which costs approximately \$0.83 on the API. The output videos are restricted to around one minute to ensure fair comparisons. Further details can be found in the \textit{Supplementary Materials}.

%----------------------------------------------------------------------------------

\subsection{Comparisons on Video Trimming}

\begin{table}[t]
\centering
\renewcommand\tabcolsep{3pt}
    \centering
    \resizebox{\linewidth}{!}{%
    \begin{tabular}{@{}c|lccccc@{}}
        \toprule
        \tb{Dataset} & \tb{Method} & Richness & Appeal & Excitement & Wasted & \tb{Average} \\
        \midrule
        \multirow{4}{*}{\rotatebox{90}{YouTube}} & {UMT}~\cite{liu2022umt} & $2.70$ & $3.08$ & $3.40$ & $3.44$ & $3.15$ \\
        &{UniVTG}~\cite{lin2023univtg}    & $2.67$ & $3.06$ & $3.35$ & $3.39$ & $3.12$ \\
        &{UVCOM}~\cite{xiao2024bridging}  & $2.72$ & $3.10$ & $3.45$ & $\mb{3.45}$ & $3.18$ \\
        &AVT (ours)                       & $\mb{2.79}$ & $\mb{3.17}$ & $\mb{3.53}$ & $3.44$ & $\mb{3.23}$ \\
        \midrule
        \multirow{4}{*}{\rotatebox{90}{TVSum}}&PGL-SUM~\cite{apostolidis2021combining} & $2.75$ & $3.05$ & $3.10$ & $3.10$ & $3.00$ \\
        &{UniVTG}~\cite{lin2023univtg}           & $2.65$ & $2.95$ & $2.85$ & $3.15$ & $2.90$ \\
        &{UVCOM}~\cite{xiao2024bridging}         & $2.50$ & $2.80$ & $2.70$ & $3.30$ & $2.83$ \\
        &AVT (ours)                              & $\mb{3.15}$ & $\mb{3.35}$ & $\mb{3.25}$ & $\mb{3.70}$ & $\mb{3.36}$ \\
        \bottomrule
    \end{tabular}}
    \caption{Evaluation agent performance of different methods on the validation set of YouTube Highlights and TVSum dataset.}
    \label{tab_gpt_eval_agent_ythl_tvsum}
\end{table}

\begin{table*}[t]
\centering
\renewcommand\tabcolsep{2pt}
\begin{minipage}{0.411\textwidth}
    \centering
    \resizebox{\textwidth}{!}{%
    \begin{tabular}{@{}lcccccccc@{}}
        \toprule
        \tb{Method} & Sup & {Dog} & {Gym.} & {Par.} & {Ska.} & {Ski.} & {Sur.} & \tb{Avg.} \\
        \midrule
        % \g{GIFs~\cite{gygli2016video2gif}}              & \g{FS} & \g{$30.8$} & \g{$33.5$} & \g{$54.0$} & \g{$55.4$} & \g{$32.8$} & \g{$54.1$} & \g{$43.4$} \\
        \g{LSVM~\cite{sun2014ranking}}                  & \g{FS} & \g{$60.0$} & \g{$41.0$} & \g{$61.0$} & \g{$62.0$} & \g{$36.0$} & \g{$61.0$} & \g{$53.6$} \\
        \g{Trailer~\cite{wang2020trailer}}              & \g{FS} & \g{$63.3$} & \g{$82.5$} & \g{$62.3$} & \g{$52.9$} & \g{$74.5$} & \g{$79.3$} & \g{$69.1$} \\
        \g{SL-Module~\cite{xu2021cross}}                & \g{FS} & \g{$70.8$} & \g{$53.2$} & \g{$77.2$} & \g{$72.5$} & \g{$66.1$} & \g{$76.2$} & \g{$69.3$} \\
        % \g{MINI-Net~\cite{hong2020mini}}              & \g{WS} & \g{$53.7$} & \g{$52.8$} & \g{$68.9$} & \g{$70.9$} & \g{$58.3$} & \g{$63.8$} & \g{$61.4$} \\
        % \g{Joint-VA~\cite{badamdorj2021joint}}        & \g{FS} & \g{$64.9$} & \g{$71.5$} & \g{$76.6$} & \g{$60.6$} & \g{$71.2$} & \g{$78.2$} & \g{$70.5$} \\
        \g{Joint-VA$\dagger$~\cite{badamdorj2021joint}} & \g{FS} & \g{$64.5$} & \g{$71.9$} & \g{$80.8$} & \g{$62.0$} & \g{$73.2$} & \g{$78.3$} & \g{$71.8$} \\
        \g{{UMT}$\dagger$~\cite{liu2022umt}}            & \g{FS} & \g{$65.9$} & \g{$75.2$} & \g{$81.6$} & \g{$71.8$} & \g{$72.3$} & \g{$82.7$} & \g{$74.9$} \\    
        % \g{{UniVTG}~\cite{lin2023univtg}}             & \g{FS} & \g{$71.8$} & \g{$76.5$} & \g{$73.9$} & \g{$73.3$} & \g{$73.2$} & \g{$82.2$} & \g{$75.2$} \\
        % w/ PT    
        \g{{UniVTG}~\cite{lin2023univtg}}               & \g{FS} & \g{$74.3$} & \g{$79.0$} & \g{$74.4$} & \g{$84.9$} & \g{$75.1$} & \g{$83.9$} & \g{$78.6$} \\
        % \g{{UniVTG}~\cite{lin2023univtg} ZS}          & \g{ZS} & \g{$36.8$} & \g{$62.8$} & \g{$65.9$} & \g{$39.2$} & \g{$64.5$} & \g{$54.0$} & \g{$53.9$} \\
        % \g{{UVCOM$^{1}$}~\cite{xiao2024bridging}}    
        \g{{UVCOM}~\cite{xiao2024bridging}}             & \g{FS} & \g{$73.8$} & \g{$77.1$} & \g{$75.7$} & \g{$75.3$} & \g{$74.0$} & \g{$82.7$} & \g{$76.4$} \\
        % \g{{UVCOM$^{2}$}~\cite{xiao2024bridging}}     & \g{FS} & \g{$66.5$} & \g{$77.4$} & \g{$82.8$} & \g{$78.7$} & \g{$74.2$} & \g{$84.6$} & \g{$77.4$} \\
        \midrule    
        \g{LIM-S~\cite{xiong2019less}}                  & \g{WS} & \g{$57.9$} & \g{$41.7$} & \g{$67.0$} & \g{$57.8$} & \g{$48.6$} & \g{$65.1$} & \g{$56.4$} \\
        \g{MINI-Net$\dagger$~\cite{hong2020mini}}       & \g{WS} & \g{$58.2$} & \g{$61.7$} & \g{$70.2$} & \g{$72.2$} & \g{$58.7$} & \g{$65.1$} & \g{$64.4$} \\
        \g{TCG$\dagger$~\cite{ye2021temporal}}          & \g{WS} & \g{$55.4$} & \g{$62.7$} & \g{$70.9$} & \g{$69.1$} & \g{$60.1$} & \g{$59.8$} & \g{$63.0$} \\ 
        \midrule
        RRAE~\cite{yang2015unsupervised}                & {ZS}   & $49.0$     & $35.0$     & $50.0$     & $25.0$     & $22.0$     & $49.0$     & $38.3$     \\
        {UniVTG}~\cite{lin2023univtg}                   & {ZS}   & $48.8$     & $57.5$     & $59.4$     & $39.7$     & $57.4$     & $49.1$     & $52.0$     \\
        {UVCOM}~\cite{xiao2024bridging}                 & {ZS}   & $46.6$     & $\mb{67.4}$& $61.4$     & $\mb{57.2}$& $63.5$     & $60.9$     & $59.5$     \\
        % AVT (ours)                                    & {ZS}   & $45.8$     & $64.0$     & $66.3$     & $25.1$     & $66.4$     & $60.6$     & $54.7$     \\
        AVT (ours)                                      & {ZS}   & $\mb{58.0}$& $62.1$     & $\mb{76.1}$& $32.0$     & $\mb{67.1}$& $\mb{67.9}$& $\mb{60.5}$\\
        \bottomrule
    \end{tabular}}
    \caption{
    {Highlight detection results of mAP on YouTube Highlights. $\dagger$ denotes using audio modality.}
    }
    \label{tab_hd_ythd}
\end{minipage}%
% \hspace{0.05\textwidth}
\hfill
\begin{minipage}{0.5648\textwidth}
    \centering
    \resizebox{\textwidth}{!}{%
    \begin{tabular}{@{}lcccccccccccccc@{}}
        \toprule
        \tb{Method} & Sup &  {VT} & {VU} & {GA} & {MS} & {PK} & {PR} & {FM} & {BK} & {BT} & {DS} & \tb{Avg.} \\
        \midrule
        \g{sLSTM~\cite{zhang2016video}}                 & \g{FS} & \g{$41.1$} & \g{$46.2$} & \g{$46.3$} & \g{$47.7$} & \g{$44.8$} & \g{$46.1$} & \g{$45.2$} & \g{$40.6$} & \g{$47.1$} & \g{$45.5$} & \g{$45.1$} \\
        \g{Trailer~\cite{wang2020trailer}}              & \g{FS} & \g{$61.3$} & \g{$54.6$} & \g{$65.7$} & \g{$60.8$} & \g{$59.1$} & \g{$70.1$} & \g{$58.2$} & \g{$64.7$} & \g{$65.6$} & \g{$68.1$} & \g{$62.8$} \\
        \g{SL-Module~\cite{xu2021cross}}                & \g{FS} & \g{$86.5$} & \g{$68.7$} & \g{$74.9$} & \g{$86.2$} & \g{$79.0$} & \g{$63.2$} & \g{$58.9$} & \g{$72.6$} & \g{$78.9$} & \g{$64.0$} & \g{$73.3$} \\
        % MINI-Net~\cite{hong2020mini}                  & {WS}   & $80.3$     & $65.3$     & $75.4$     & $81.3$     & $78.0$     & $54.5$     & $55.9$     & $71.7$     & $76.9$     & $59.1$     & $69.8$     \\
        % Joint-VA~\cite{badamdorj2021joint}            & {FS}   & $83.4$     & $64.7$     & $84.4$     & $86.5$     & $70.3$     & $67.5$     & $66.9$     & $68.1$     & $95.0$     & $60.8$     & $74.8$     \\
        \g{Joint-VA$\dagger$~\cite{badamdorj2021joint}} & \g{FS} & \g{$83.7$} & \g{$57.3$} & \g{$78.5$} & \g{$86.1$} & \g{$80.1$} & \g{$69.2$} & \g{$70.0$} & \g{$73.0$} & \g{$97.4$} & \g{$67.5$} & \g{$76.3$} \\
        \g{UMT$\dagger$~\cite{liu2022umt}}              & \g{FS} & \g{$87.5$} & \g{$81.5$} & \g{$88.2$} & \g{$78.8$} & \g{$81.5$} & \g{$87.0$} & \g{$76.0$} & \g{$86.9$} & \g{$84.4$} & \g{$79.6$} & \g{$83.1$} \\
        % {UniVTG}~\cite{lin2023univtg}                 & {FS}   & $83.9$     & $85.1$     & ${89.0}$   & ${80.1}$   & ${84.6}$   & ${81.4}$   & ${70.9}$   & ${91.7}$   & ${73.5}$   & ${69.3}$   & ${81.0}$   \\
        % w/ PT
        \g{UniVTG~\cite{lin2023univtg}}                 & \g{FS} & \g{$92.0$} & \g{$77.8$} & \g{$89.8$} & \g{$83.8$} & \g{$82.2$} & \g{$85.8$} & \g{$74.3$} & \g{$91.8$} & \g{$90.5$} & \g{$77.6$} & \g{$84.6$} \\
        % {UniVTG}~\cite{lin2023univtg} ZS              & {ZS}   &  ${78.5}$  & ${67.0}$   & ${75.3}$   & ${63.6}$   & ${67.0}$   & ${66.8}$   & ${35.4}$   & ${85.3}$   & ${83.1}$   & ${50.0}$   & ${67.2}$   \\
        % *** \g{QD-DETR~\cite{moon2023query}}                & \g{FS} & \g{$88.2$} & \g{$87.4$} & \g{$85.6$} & \g{$85.0$} & \g{$85.8$} & \g{$86.9$} & \g{$76.4$} & \g{$91.3$} & \g{$89.2$} & \g{$73.7$} & \g{$85.0$} \\
        % \midrule        
        \g{UVCOM~\cite{xiao2024bridging}}               & \g{FS} & \g{$87.6$} & \g{$91.6$} & \g{$91.4$} & \g{$86.7$} & \g{$86.9$} & \g{$86.9$} & \g{$76.9$} & \g{$92.3$} & \g{$87.4$} & \g{$75.6$} & \g{$86.3$} \\
        \midrule     
        \g{LIM-S~\cite{xiong2019less}}                  & \g{WS} & \g{$55.9$} & \g{$42.9$} & \g{$61.2$} & \g{$54.0$} & \g{$60.4$} & \g{$47.5$} & \g{$43.2$} & \g{$66.3$} & \g{$69.1$} & \g{$62.6$} & \g{$56.3$} \\
        \g{MINI-Net$\dagger$~\cite{hong2020mini}}       & \g{WS} & \g{$80.6$} & \g{$68.3$} & \g{$78.2$} & \g{$81.8$} & \g{$78.1$} & \g{$65.8$} & \g{$57.8$} & \g{$75.0$} & \g{$80.2$} & \g{$65.5$} & \g{$73.2$} \\
        \g{TCG$\dagger$~\cite{ye2021temporal}}          & \g{WS} & \g{$85.0$} & \g{$71.4$} & \g{$81.9$} & \g{$78.6$} & \g{$80.2$} & \g{$75.5$} & \g{$71.6$} & \g{$77.3$} & \g{$78.6$} & \g{$68.1$} & \g{$76.8$} \\
        \midrule     
        SG~\cite{mahasseni2017unsupervised}             & {ZS}   & $42.3$     & $47.2$     & $47.5$     & $48.9$     & $45.6$     & $47.3$     & $46.4$     & $41.7$     & $48.3$     & $46.6$     & $46.2$     \\
        {UniVTG}~\cite{lin2023univtg}                   & {ZS}   & $52.0$     & $48.1$     & $50.9$     & $56.9$     & $51.6$     & $43.3$     & $\mb{60.0}$& $\mb{64.0}$& $\mb{59.2}$& $54.9$     & $54.1$     \\
        {UVCOM}~\cite{xiao2024bridging}                 & {ZS}   & $63.4$     & $44.5$     & $50.6$     & $\mb{67.6}$& $55.1$     & $42.0$     & $47.5$     & $56.9$     & $58.6$     & $39.3$     & $52.5$     \\
        AVT (ours)                                      & {ZS}   & $\mb{76.6}$& $\mb{75.9}$& $\mb{62.4}$& ${63.9}$   & $\mb{76.6}$& $\mb{68.8}$& $39.4$     & $45.6$     & $43.4$     & $\mb{62.9}$& $\mb{61.6}$\\
        \bottomrule
    \end{tabular}}
    \caption{Highlight detection results of Top-5 mAP on TVSum. $\dagger$ denotes using audio modality.
    \tb{FS}: Fully supervised.
    \tb{WS}: Weakly supervised.
    \tb{ZS}: Zero-shot.}
    \label{tab_hd_tvsum}
\end{minipage}
\end{table*}

%----------------------------------------------------------------------------------

\noindent\textbf{Human Evaluation.} We conduct a user study based on our constructed video trimming dataset. We design a blind test by randomly shuffling the order of output videos from each method.
For the existing highlight detection methods, we utilize their pretrained models to generate saliency scores and obtain final videos by concatenating the video intervals with top scores.
Seventeen participants are asked to score these videos across five aspects, similar to the criteria of the Video Evaluation Agent: Material Richness, Appeal, Exciting Segments, Amount of Wasted Footage, and an overall perception score.
{Tab.}~\ref{tab_blind_test_avt} shows that our AVT achieves overall improvements in the quality of the final cut, benefiting from the filtering of wasted footage and the clip composition process. We also display the agent evaluation score in the rightmost column, which is consistent with the ranking from human evaluation.

%----------------------------------------------------------------------------------

\noindent\textbf{Agent Evaluation.} Following Sec.~\ref{sec_video_evaluation_agent}, we evaluate the quality of the generated video using our designed Video Evaluation Agent. We conduct experiments on the validation sets of the YouTube Highlights and TVSum datasets, using 150 and 10 videos, respectively. We compare our approach with three related previous methods. 
{Tab.}~\ref{tab_gpt_eval_agent_ythl_tvsum} show that our AVT achieves higher metrics than existing methods. Notably, the ablation study on the consistency with human evaluation is further elaborated in Sec.~\ref{sec_ablation}.

\subsection{Comparisons on Highlight Detection}
In this section, we compare our method with previous highlight detection and video summarization approaches. We denote the saliency score of AVT as follows:
\begin{equation}
S_i = \left\{
\begin{array}{ll}
S_{h} & \text{if } S_{h} > max(S_{d}), \\
S_{h} - max(S_{d}) & \text{otherwise},
\end{array}
\right.
\end{equation}
where $i$ represents the $Clip \ ID$, and $S_{h}$ and $S_{d}$ denote the highlight and defect scores in the structured clip description.
We first conduct experiments on YouTube Highlights and TVSum. We report the results from the original papers of the fully/weakly supervised methods. 
Then, we compare the methods in terms of zero-shot transfer. 
Notably, for UniVTG~\cite{lin2023univtg}, we directly copy their zero-shot result. 
For others, we infer with their models pretrained on the largest scaled datasets, such as QVHighlights~\cite{lei2021qvhighlights} and Charades-STA~\cite{gao2017tall}.
We follow the validation splits of~\cite{lin2023univtg,liu2022umt} to compare on the YouTube Highlights dataset. As the scale of TVSum is small, with its validation set containing only 10 videos, which may yield inconsistent scores, we measure all videos up to 50 for the zero-shot settings. 
As shown in {Tab.}~\ref{tab_hd_ythd} and {Tab.}~\ref{tab_hd_tvsum}, our AVT achieves state-of-the-art highlight detection performance under zero-shot transfer and is comparable to partially supervised methods with training.

\begin{figure}[t]
    \centering
    \includegraphics[width=\linewidth]{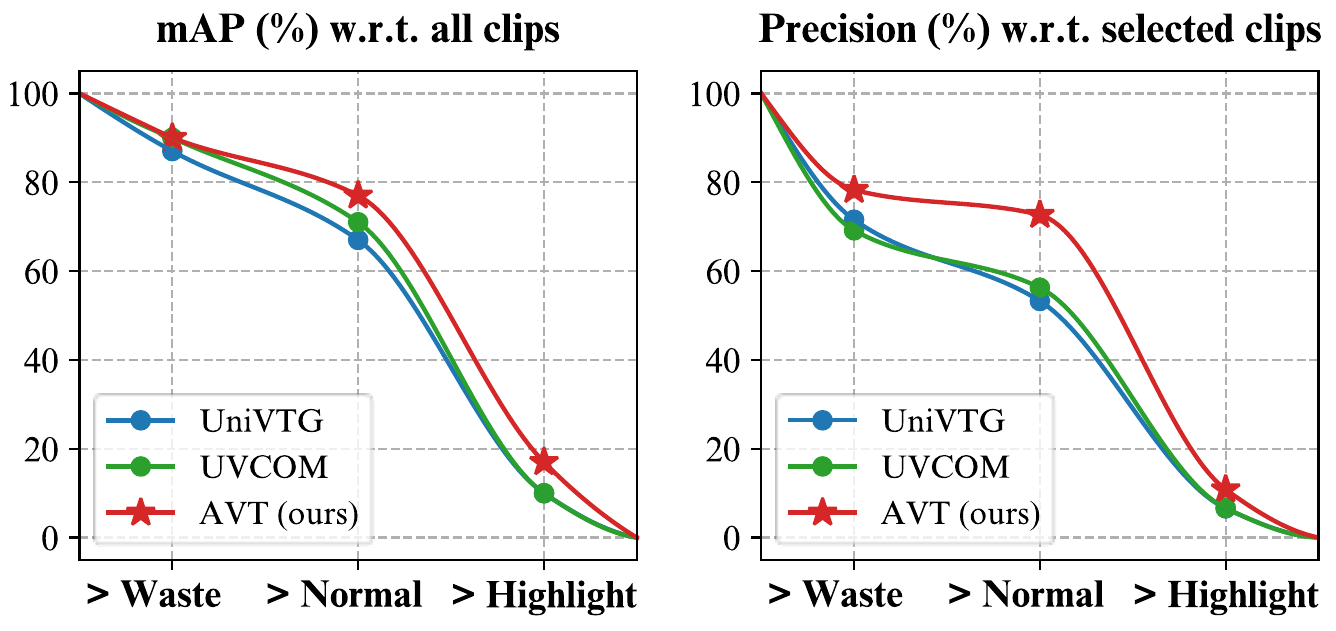}
    \caption{
    {Highlight detection results of mAP and precision on our collected video trimming dataset.}
    }
    \label{fig_avt_hd_and_waste}
\end{figure}

Next, using our constructed video trimming dataset, we show the mAP of highlight detection with existing methods and the precision of the highlight segments in the selected clips for the final video. In Fig.~\ref{fig_avt_hd_and_waste}, we observe that there is less wasted footage selected in AVT videos than in previous methods, as they do not focus on footage filtering. Additionally, our method derives more highlight segments.

\begin{figure*}[t]
    \centering
    \includegraphics[width=\textwidth]{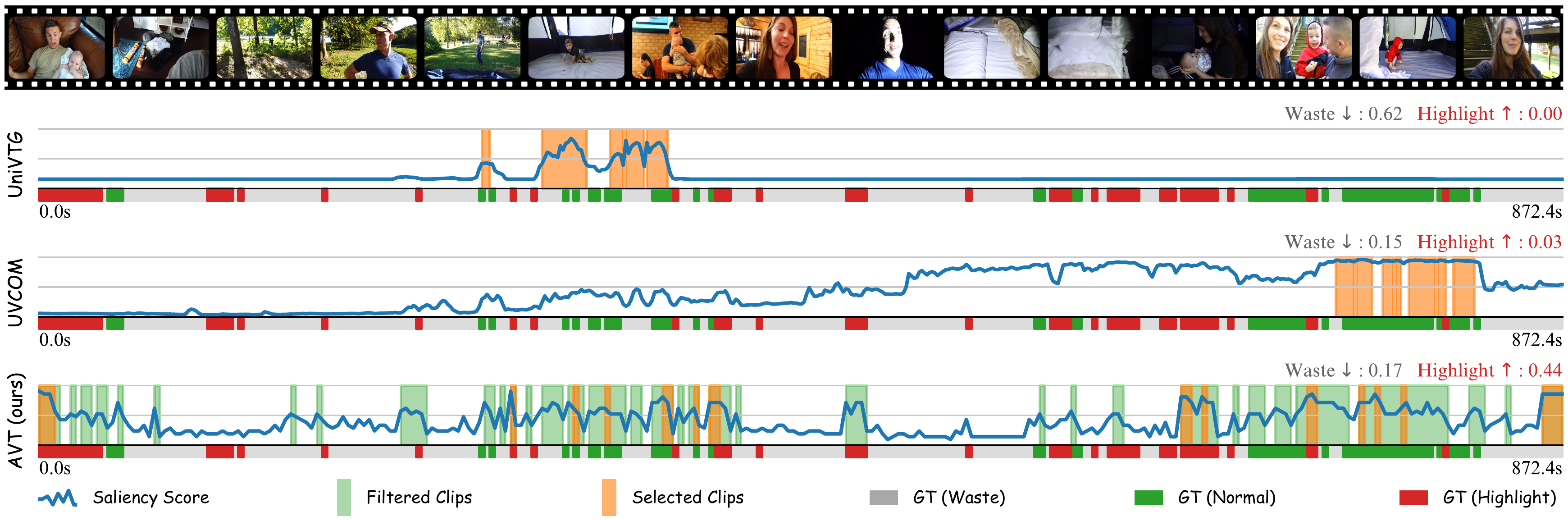}
    \caption{
    Visualization of trimmed videos on the video trimming dataset. AVT creates a more complete storyline with more highlight footage and less wasted footage.
    }
    \label{fig_visualization}
\end{figure*}

\begin{table}
    \renewcommand\tabcolsep{4pt}
    \centering
    \resizebox{\linewidth}{!}{%
    \begin{tabular}{@{}lccccccc@{}}
        \toprule
        \tb{Method} & VS & CF & DF & SC & \tb{User} $\uparrow$ & \tb{Waste} $\downarrow$ & \tb{Highlight} $\uparrow$ \\
        \midrule
        {UniVTG}~\cite{lin2023univtg}   & \No  & \No  & \No  & \No  & $6.30$ & $0.276$ & $0.066$ \\
        {UVCOM}~\cite{xiao2024bridging} & \No  & \No  & \No  & \No  & $6.23$ & $0.175$ & $0.066$ \\         
        \midrule    
        \multirow{6}{*}{AVT (ours)}     & \No  & \No  & \No  & \No  & $3.70$ & $0.337$ & $0.083$ \\
                                        & \Yes & \No  & \No  & \No  & $6.19$ & $0.135$ & $\mb{0.110}$ \\
                                        & \Yes & \Yes & \No  & \No  & $6.45$ & $0.165$ & $0.096$ \\
                                        & \Yes & \Yes & \Yes & \No  & $6.70$ & $0.141$ & $0.109$ \\
                                        & \Yes & \No  & \No  & \Yes & $5.23$ & $0.199$ & $0.107$ \\
                                        & \Yes & \Yes & \Yes & \Yes & $\mb{7.15}$ & $\mb{0.083}$ & $0.108$ \\
        \bottomrule
    \end{tabular}
    }
    \caption{
    Ablation study on the effectiveness of AVT components.
    \textbf{VS}: Video Structuring.
    \textbf{CF}: Clip Filtering.
    \textbf{DF}: Dynamic Filter.
    \textbf{SC}: Story Composition.
    }
    \label{tab_components}
\end{table}

\subsection{Ablation Study}\label{sec_ablation}

\noindent\textbf{Components of AVT.} In this section, we analyze the effectiveness of each component within AVT, including the structuring phase, filtering phase, and dynamic module. Additionally, we compare the results of replacing the story composition phase with a simple concatenation of video slices in temporal order. For all experiments without the composition phase, clips with the top saliency scores are selected. For the control condition, with all components disabled, we randomly select video clips. We conducted a user study and quantitative precision measurements on the waste/highlight footage ratios to assess the quality of the generated videos. {Tab.}~\ref{tab_components} shows that clip filtering significantly reduces wasted footage, while the composition process enhances overall video impressions. Notably, without the dynamic filter module, particularly for sports content, highlight segments may be discarded by the defect attribute. Further analysis is in the \textit{Supplementary Materials}.

\noindent\textbf{Human Correlation of Evaluation Agent.}
Following G-Eval~\cite{liu2023g}, we adopt three meta-evaluation metrics: Pearson ($r$), Spearman ($\rho$), and Kendall-Tau ($\tau$), to measure the correlation between our evaluation agent and human preferences. We perform ablation studies on the prompt settings to investigate the impact of requesting the agent to provide reasons alongside the score, as well as the effect of using the diverse criteria outlined in Sec.~\ref{sec_video_evaluation_agent}. With both strategies activated, we achieve an average correlation of $0.5247$ with human ratings, as shown in {Tab.}~\ref{tab_correlations}.

\begin{table}
    \renewcommand\tabcolsep{2pt}
    \centering
    \resizebox{\linewidth}{!}{%
    \begin{tabular}{@{}cccccc@{}}
        \toprule
        Output Reason & Diverse Criteria & $\bm{r}$ & $\bm{\rho}$ & $\bm{\tau}$ & \tb{Avg.} \\
        \midrule
        \No  & \No  & $0.2675$      & $0.2451$      & $0.1723$      & $0.2283$ \\
        \No  & \Yes & $0.4082$      & $0.4119$      & $0.3067$      & $0.3756$ \\
        \Yes & \No  & $0.5260$      & $0.4990$      & $0.3738$      & $0.4663$ \\
        \Yes & \Yes & $\mb{0.5616}$ & $\mb{0.5667}$ & $\mb{0.4457}$ & $\mb{0.5247}$ \\
        \bottomrule
    \end{tabular}
    }%
    \caption{
    Pearson ($r$), Spearman ($\rho$), and Kendall-Tau ($\tau$) correlations of different metrics on video trimming benchmark.
    }
    \label{tab_correlations}
\end{table}

\subsection{Visualization}
We visualize the saliency scores and the selected intervals of each method in Fig.~\ref{fig_visualization}. Since existing methods do not include a composition phase, their final video is constructed by concatenating high-salient footage, resulting in an inconsistent viewing experience and limited content richness. AVT surpasses previous works by selecting more highlight footage and less wasted footage while maintaining a consistent storyline with the raw videos.

%-------------------------------------------------------------------------

\section{Conclusion}
In this paper, we introduce the novel task of Video Trimming (VT), which focuses on segment selection and narrative preservation to extract meaningful insights from redundant content.
To tackle this task, we propose Agent-based Video Trimming (AVT), a baseline framework with three key phases: Video Structuring, where a Video Captioning Agent provides segment descriptions; Clip Filtering, which dynamically selects clips using a filtering module; and Story Composition, where a Video Arrangement Agent creates a cohesive narrative.
Further, a Video Evaluation Agent is designed to assess video quality. We construct a benchmark annotated for video trimming tasks. Our approach outperforms existing methods in highlight detection and demonstrates superior human preference in user studies.

{
    \small
    \bibliographystyle{ieeenat_fullname}
    \bibliography{main}
}

%------------------------------------------------------------------------------------------------------------------------------------------------------------------------------------------------------------------------------------------------------------------------------

% \iffalse
\newcommand{\chinese}[1]{\begin{CJK*}{UTF8}{gbsn}#1\end{CJK*}}
\newcommand{\korean}[1]{\begin{CJK*}{UTF8}{}\CJKfamily{mj}#1\end{CJK*}}
\renewcommand{\thetable}{S\arabic{table}}
\renewcommand{\thefigure}{S\arabic{figure}}
\renewcommand{\theequation}{S\arabic{equation}}
\setcounter{table}{0}
\setcounter{figure}{0}
\setcounter{equation}{0}

\maketitlesupplementary
\appendix

\begin{table}[ht]
\centering
    \renewcommand{\arraystretch}{1.33}
    \renewcommand\tabcolsep{4pt}
    \resizebox{\linewidth}{!}{%
    \begin{tabular}{l|l|l|c}
        \toprule
        \textbf{Class}               & \textbf{Sub-Class}           & \textbf{User}               & \textbf{YouTube ID(s)}             \\
        \midrule    
        \multirow{10}{*}{daily life} & \multirow{4}{*}{family}      & CapperCoolCooper            & \texttt{KRqR6eLSoP8}               \\
                                     &                              & Earls Family Vlogs          & \texttt{MyLwV1V19WY}               \\
                                     &                              & Jason Boon                  & \texttt{PcxuFef17PY}               \\
                                     &                              & The Semps                   & \texttt{YoIkzzpQjKM}               \\
        \cline{2-4} 
                                     & \multirow{1}{*}{food}        & Soon Films \korean{순필름} & \texttt{\_McjBGfacfc}               \\
        \cline{2-4} 
                                     & \multirow{1}{*}{friend}      & Shawn Roscoe                & \texttt{J0nA4VgnoCo}               \\
        \cline{2-4}
                                     & \multirow{1}{*}{light show}  & World In Nature \chinese{自然} & \texttt{nFPJMj0tq9G}               \\
        \cline{2-4}
                                     & \multirow{3}{*}{pets}        & Cats with GoPro             & \texttt{a98Ra7PaTeE}               \\
                                     &                              & Gone to the Snow Dogs       & \texttt{iISfWRiDe1g}               \\
                                     &                              & Ms Kendall G                & \texttt{Bhxk-O1Y7Ho}               \\
        \midrule
        \multirow{21}{*}{sports}     & \multirow{1}{*}{badminton}   & SHUTTLE\&MORE               & \texttt{lyhfZy7tShU}               \\
        \cline{2-4}
                                     & \multirow{1}{*}{basketball}  & June young Kim              & \texttt{UNemjRp6YJg}               \\
        \cline{2-4}
                                     & \multirow{4}{*}{cycling}     & \multirow{3}{*}{Erkan Sakallioglu} & \texttt{AO6p0jlOd6U}             \\
                                     &                              &                             & \texttt{iwCSAYGwPq4}               \\
                                     &                              &                             & \texttt{xLhHU8uo2aY}               \\
        \cline{3-4}
                                     &                              & Richard Whittle             & \texttt{g-\_-B5HBRlY}              \\
        \cline{2-4}
                                     & \multirow{1}{*}{motorcycle}  & Skaily Production           & \texttt{MXAzBe7PZOQ}               \\
        \cline{2-4}
                                     & \multirow{1}{*}{skateboard}  & TOW TRUCK BOB               & \texttt{VVK1KkIKCYQ}               \\
        \cline{2-4}
                                     & \multirow{3}{*}{skating}     & \multirow{3}{*}{HC+}        & \texttt{a8M-5nTrll8}               \\
                                     &                              &                             & \texttt{bA3CxZllhsI}               \\
                                     &                              &                             & \texttt{dZ3i-HuhQXM}               \\
        \cline{2-4} 
                                     & \multirow{10}{*}{skiing}     & \multirow{8}{*}{Alex E}     & \texttt{E5OqGoDzNtg}               \\
                                     &                              &                             & \texttt{dFrfsgW1M98}               \\
                                     &                              &                             & \texttt{ddxw58h5Y\_A}              \\
                                     &                              &                             & \texttt{fB8zm1hTvgA}               \\
                                     &                              &                             & \texttt{h7aeRrf-m-8}               \\
                                     &                              &                             & \texttt{mqNZjVDZcfY}               \\
                                     &                              &                             & \texttt{puAxGH6aWMY}               \\
                                     &                              &                             & \texttt{xnsTcvtttfY}               \\
        \cline{3-4} 
                                     &                              & Emerson Nishi               & \texttt{WL4TA--CVcA}               \\
        \cline{3-4} 
                                     &                              & ecosnowsportsTV             & \texttt{5FE87lIj1DQ}               \\
        \midrule
        \multirow{11}{*}{travel}     & \multirow{1}{*}{amusement park} & Informative Topics Vlogs    & \texttt{hmImxd681YI}               \\
        \cline{2-4}
                                     & \multirow{1}{*}{city walk}   & Jahaar views                & \texttt{tp912d19x4E}               \\
        \cline{2-4}
                                     & \multirow{1}{*}{hiking}      & thePOVchannel               & \texttt{I1gGZa4-h\_U}              \\
        \cline{2-4}
                                     & \multirow{1}{*}{luge}        & Travel\&Adventure Junkies    & \texttt{dXZO0qnhrPo}               \\
        \cline{2-4}
                                     & \multirow{1}{*}{mountaineering} & stivn                     & \texttt{iUMBQugUtVQ}               \\
        \cline{2-4}
                                     & \multirow{1}{*}{rafting}     & All About Shenzhen          & \texttt{A7Ys5d-Zwro}               \\
        \cline{2-4}
                                     & \multirow{1}{*}{road trip}   & Mojo Rides                  & \texttt{ePKyNYP7uNg}               \\
        \cline{2-4}
                                     & \multirow{2}{*}{show}        & HetfieldMustaine22           & \texttt{McC9gB5Cr6o}               \\
        \cline{3-4}
                                     &                              & KriyaLv                     & \texttt{1OLM0\_Jzt5M}              \\
        \cline{2-4}
                                     & \multirow{2}{*}{water park}& \multirow{2}{*}{Gezen Adam} & \texttt{3iz5SmEQj9A}               \\
                                     &                              &                             & \texttt{Wgbe-WTp\_QI}              \\
        \bottomrule
    \end{tabular}}
    \vspace{4pt}
\caption{Our curated video trimming dataset includes 42 YouTube videos, contributed by 30 users, and spans a wide variety of topics.}
\label{tab_dataset_id}
\vspace{-3pt}
\end{table}

%-------------------------------------------------------------------------

\section{Video Trimming Dataset}
\begin{figure*}
    \begin{minipage}[b]{\linewidth}
    \centering
    \renewcommand\tabcolsep{5pt}
    \resizebox{\linewidth}{!}{%
    \begin{tabular}{@{}lcccccc@{}}
        \toprule
        \textbf{Dataset} & \textbf{\#Video} & \textbf{\#User} & \textbf{Content} & \textbf{Annotation type} & \textbf{Query} & \textbf{Duration (Min, Max, Avg)} \\ 
        \midrule
        YouTube Highlights~\cite{sun2014ranking} & 423    & 5          & Web videos            & Frame-level scores & Title & 7s, 1483s, 102s                  \\
        SumMe \cite{gygli2014creating}           & 25     & 15$\sim$18 & User-generated videos & Frame-level scores & N/A   & 32s, 324s, 146s                  \\
        TVSum \cite{song2015tvsum}               & 50     & 20         & Web videos            & Frame-level scores & Title & 83s, 647s, 235s                  \\
        Charades-STA~\cite{gao2017tall}          & 9,848  & 267        & Web videos            & Time intervals     & Local caption & 2s, 194s, 30s                    \\
        OVP \cite{de2011vsumm}                   & 50     & 5          & Various genre videos  & Time intervals     & N/A   & 83s, 647s, 235s                  \\
        YouTube \cite{de2011vsumm}               & 39     & 5          & Web videos            & Time intervals     & N/A   & 83s, 647s, 235s                  \\ 
        \midrule
        Video Trimming (ours)                    & 42     & 10         & Web videos            & Frame-level scores & N/A   & \tb{141s}, \tb{1483s}, \tb{556s} \\
        \bottomrule
    \end{tabular}}
    \captionof{table}{Comparisons of existing datasets with our video trimming dataset.}
    \vspace{20pt}
    \label{tab_dataset_comparison}
\end{minipage}
\begin{minipage}[b]{\linewidth}
    \centering
    \includegraphics[width=\textwidth]{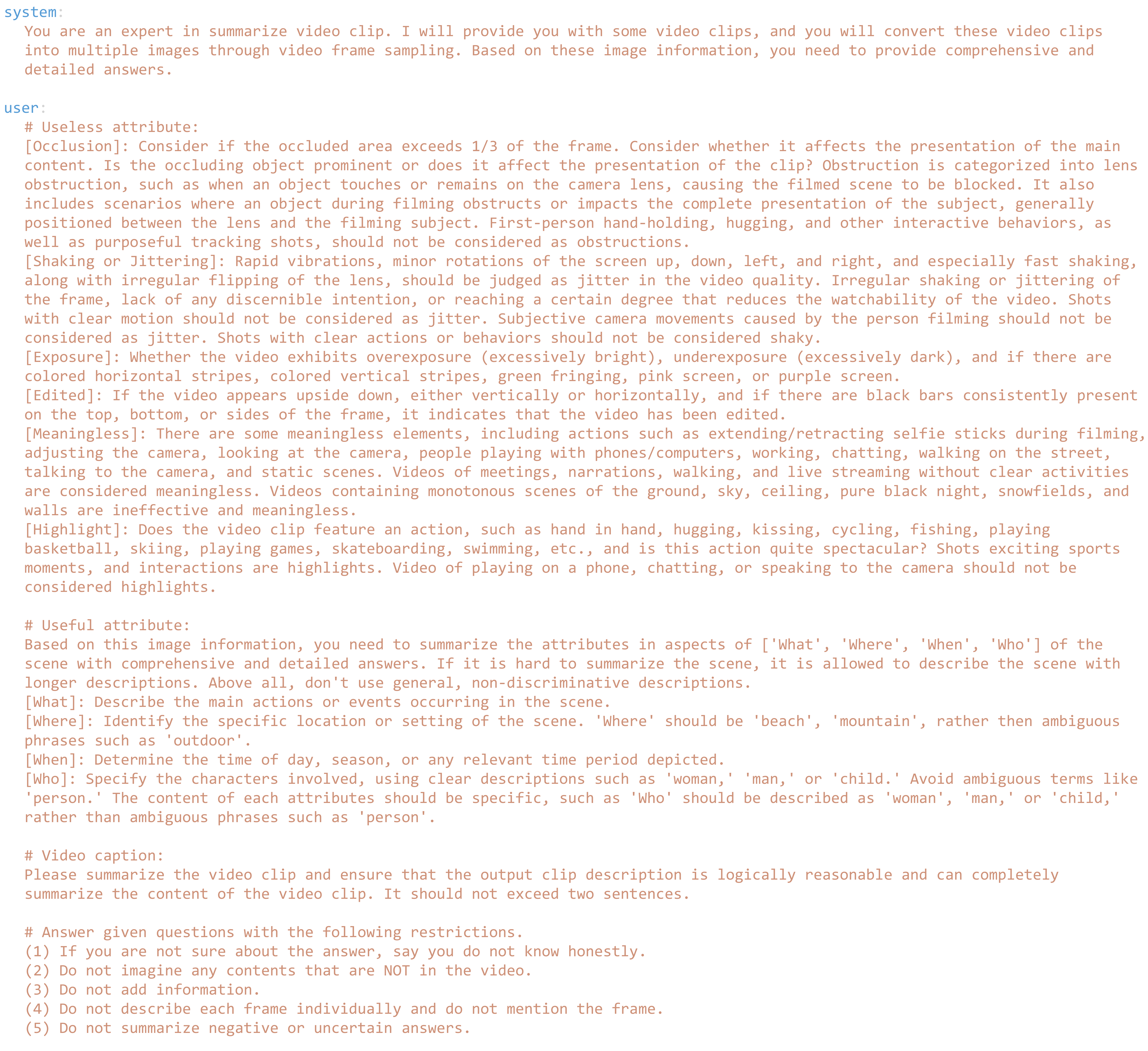}
\end{minipage}
\end{figure*}

We collect user-generated videos from YouTube and construct a benchmark for video trimming specifically. The data collection process adheres to three key principles:

\begin{itemize}
\item[1.] The current dataset predominantly consists of videos already edited by their creators. In contrast, for video trimming scenarios, we aim to use raw, unedited videos. These raw videos, typically filmed by individuals, often contain imperfections such as occlusion, jitter, or overexposure, reflecting real-world filming conditions.
\item[2.] The raw videos selected should be in long form, with durations exceeding 5 minutes, rather than short, pre-edited montages.
\item[3.] In practical video trimming tasks, input videos may originate from multiple sources. As a result, algorithms are expected to generate cuts from various videos, whereas existing datasets typically focus on a single video per topic.
\end{itemize}
Following these principles, we curated a collection of 42 videos uploaded by 30 different users. A comparison with existing datasets is shown in Tab.~\ref{tab_dataset_comparison}, which highlights that our dataset boasts the longest average video duration, approximately 10 minutes per video.

To ensure diversity in trimming scenarios, we selected videos spanning a range of topics, categorized into three groups: daily life, sports, and travel vlogs. For each category, we chose 10 video uploaders and included one or more videos that revolve around a consistent event. Detailed topics and corresponding YouTube IDs are listed in Tab.~\ref{tab_dataset_id}.

Additionally, we annotated each video with 10 annotators using four ranking levels to evaluate footage quality: 0 for wasted, 1 for ambiguous, 2 for normal, and 3 for highlight footage. Examples of these ground-truth annotations can be found in Sec~\ref{more_vis}.

\section{Prompt Design} 
\begin{figure*}
\begin{minipage}[b]{\textwidth}
    \centering
    \includegraphics[width=\textwidth]{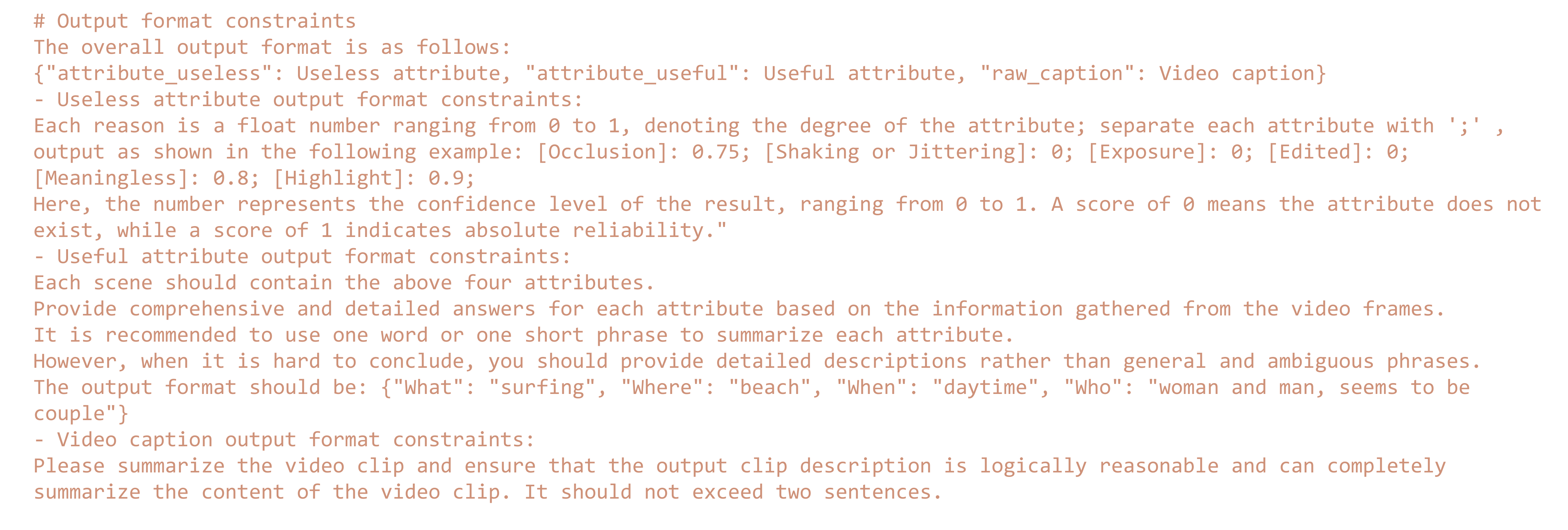}
    \caption{
    Prompt for video structuring.
    }
    \label{fig_prompt_video_structuring}
    \vspace{15pt}
\end{minipage}
\vfill
\begin{minipage}[b]{\textwidth}
    \centering
    \includegraphics[width=\textwidth]{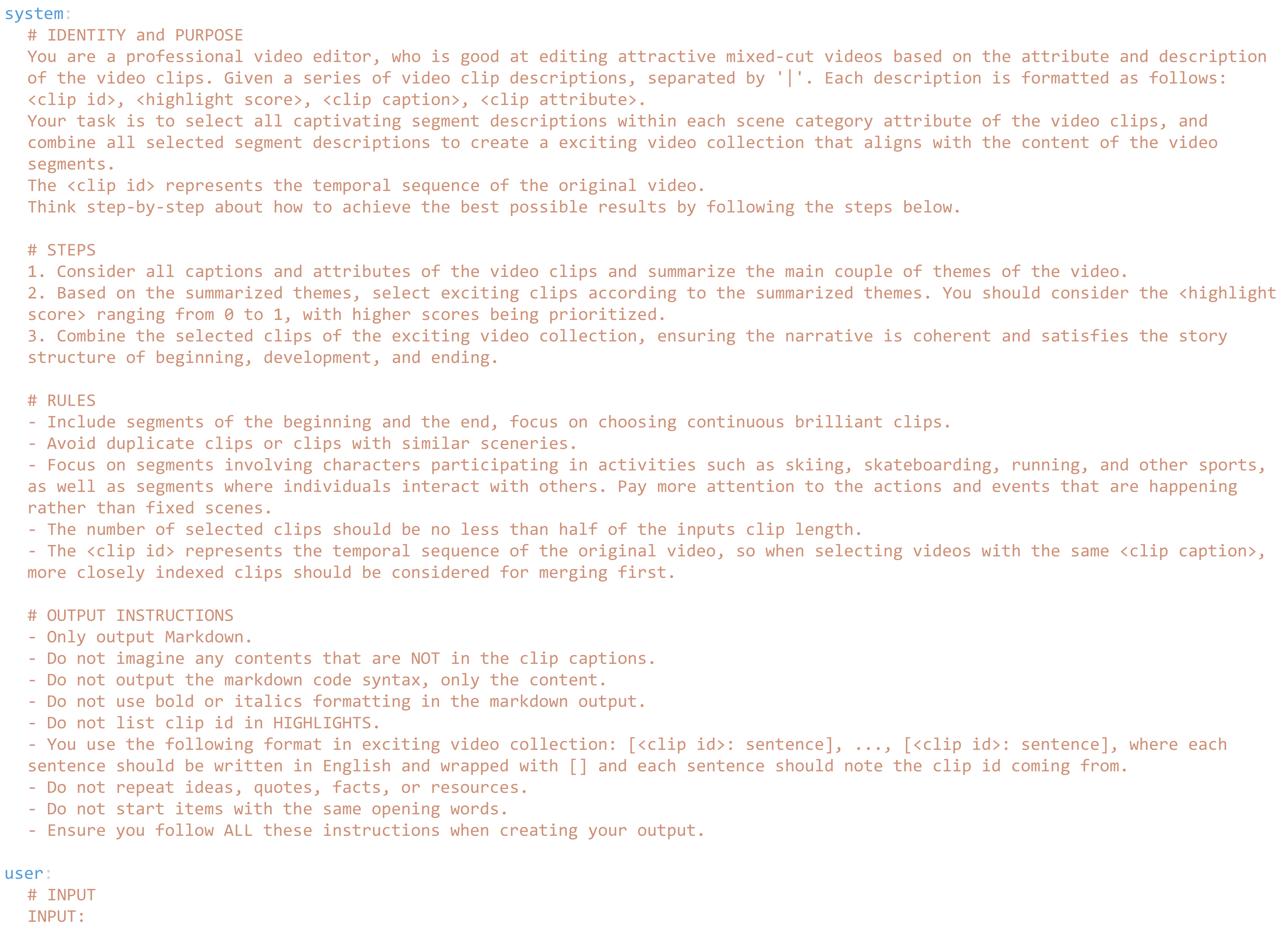}
    \caption{
    Prompt for story composition with grouped clips.
    }
    \label{fig_prompt_story_step1}
\end{minipage}
\end{figure*}

This section provides the detailed prompts for AVT, utilizing GPT-4o~\cite{gpt4o}. It covers prompts for video structuring, story composition, and video evaluation. Figure~\ref{fig_prompt_video_structuring} illustrates the structuring prompt, which includes general captions, defect attributes, and contextual attributes.

\begin{figure*}[ht]
    \centering
    \includegraphics[width=\textwidth]{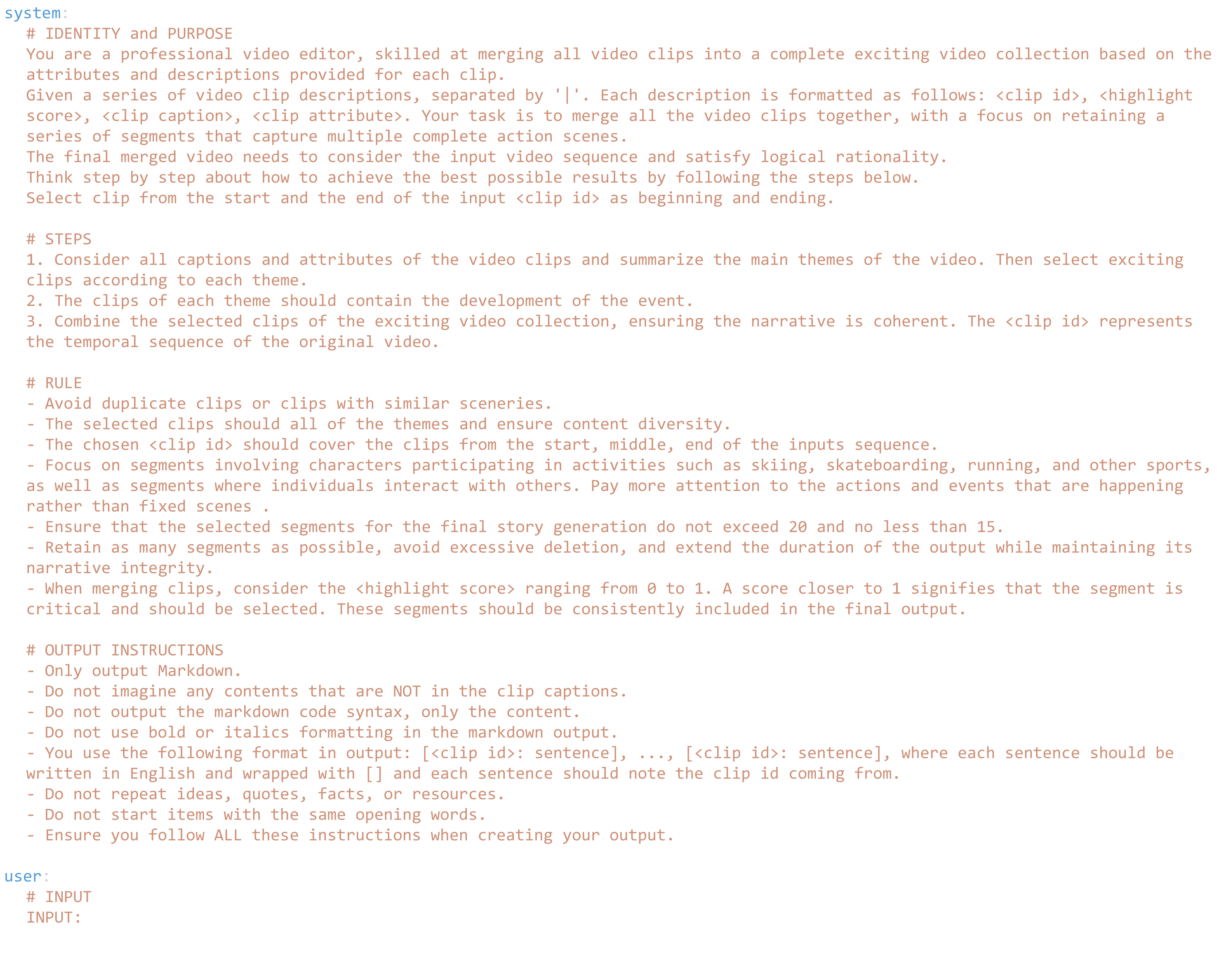}
    \caption{
    Prompt for story composition with global clips.
    }
    \label{fig_prompt_story_step2}
\end{figure*}

\begin{figure*}
\begin{minipage}[b]{\textwidth}
    \centering
    \includegraphics[width=\textwidth]{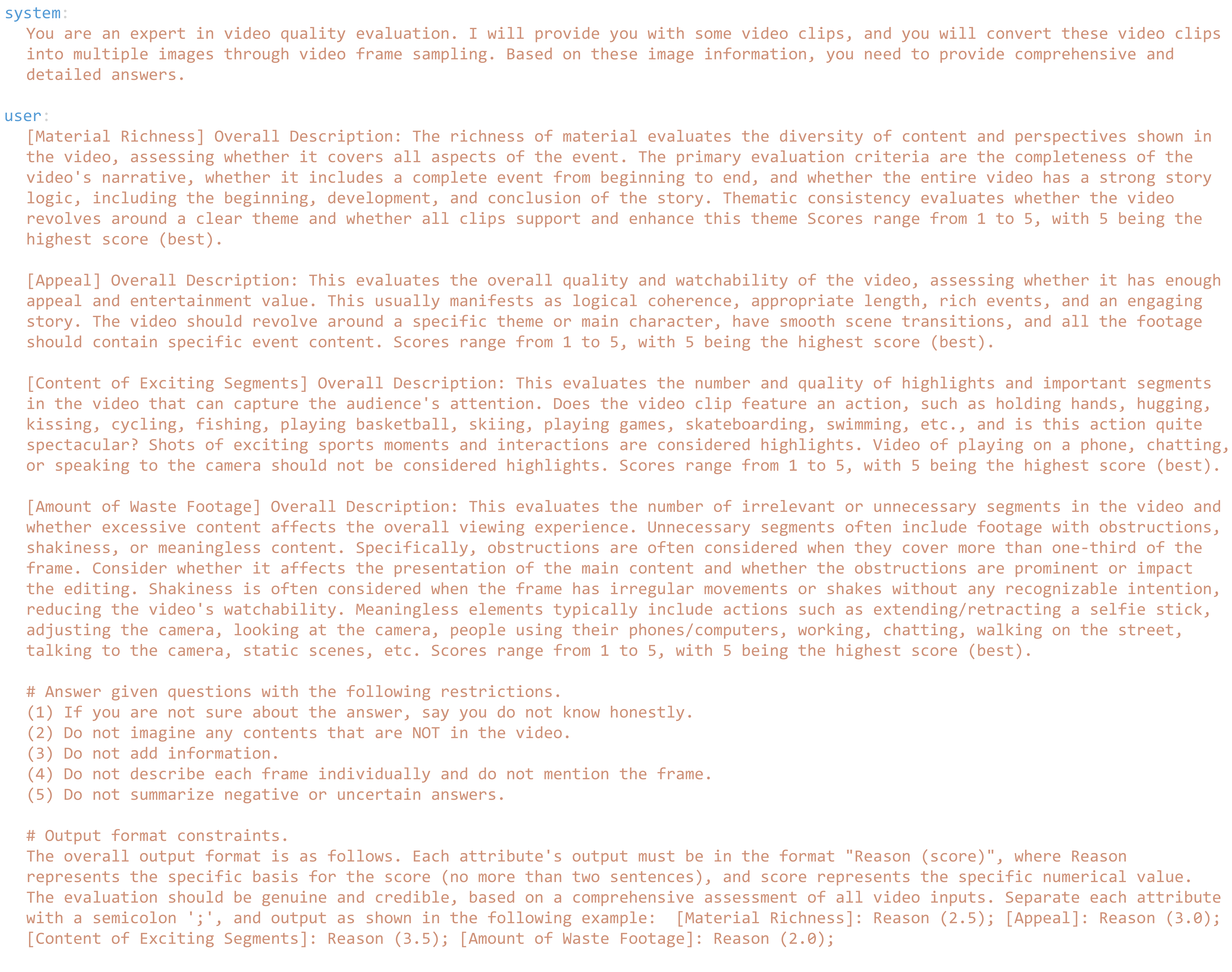}
% \vspace{-10pt}
    \caption{
    Prompt for video evaluation.
    }
    \label{fig_prompt_eval}
\vspace{30pt}
\end{minipage}
\begin{minipage}[b]{\linewidth}
    \renewcommand\tabcolsep{5pt}
    \centering
    \resizebox{\linewidth}{!}{%
    \begin{tabular}{@{}ccccccc@{}}
        \toprule
        Frame Sampling Ratio & Prompt   &Input Image Token & Input Text Token& Output Text Token& API Cost & Agent Metric \\
        \midrule
        4 / 1s & Isolated  & 1,836,000 & 100,000 & 20,000 & \$5.04  & 3.34 \\
        4 / 1s & Unified   & \ \ \ 612,000   & 100,000 & 20,000 & \$1.98  & 3.33 \\
        1 / 1s & Isolated  & \ \ \ 459,000   & 100,000 & 20,000 & \$1.60  & 3.34 \\
        1 / 1s & Unified   & \ \ \ 153,000   & 100,000 & 20,000 & \$0.83  & 3.32 \\
        \bottomrule
    \end{tabular}}
    \captionof{table}{Ablation study on the impact of sampling ratio and prompt design on performance and cost.}
    \label{tab_efficiency}
\end{minipage}
\end{figure*}

Next, we describe the story composition process, which occurs in two stages. The first stage focuses on grouping clips to prevent overwhelming input lengths, prioritizing the selection of highlight segments (Fig.~\ref{fig_prompt_story_step1}). In the second stage, the selected clips are gathered and arranged into a coherent final video, with an emphasis on narrative flow (Fig.~\ref{fig_prompt_story_step2}). Finally, Figure~\ref{fig_prompt_eval} presents the prompt used by GPT to evaluate video quality.

\clearpage

\clearpage

\clearpage

\begin{table}[ht]
\centering
\renewcommand\tabcolsep{9pt}
    \centering
    \resizebox{0.9\linewidth}{!}{%
    \begin{tabular}{@{}lccc@{}}
        \toprule
        \tb{Method} & ViCLIP & InternVideo2 & \tb{Avg.}  \\
        \midrule
        {UniVTG}~\cite{lin2023univtg}    & 0.877 & 0.941 & 0.909 \\
        {UVCOM}~\cite{xiao2024bridging}  & 0.852 & 0.928 & 0.890 \\
        AVT (ours)                       & \tb{0.906} & \tb{0.951} & \tb{0.929} \\
        \bottomrule
    \end{tabular}
    }%
    \captionof{table}{Comparison of the fidelity between the final videos and the raw videos.}
    \label{tab_viclip_similarity}
    \vspace{-2pt}
\end{table}

%-------------------------------------------------------------------------

\section{Implementation and Efficiency}
We analyze the efficiency and cost of different implementations by varying the video sampling ratio and prompt design. Specifically, we compare a sampling ratio of 1 frame per second (fps) with 4 fps. As shown in Fig.~\ref{fig_prompt_video_structuring}, three components: raw captions, defect attributes, and contextual attributes, are typically generated together using a unified prompt. Alternatively, these components can be extracted separately using isolated prompts, which require processing three times the visual content.

The current GPT API pricing is \$2.50 per million input tokens and \$10.00 per million output tokens. Each sampled keyframe resized to $512 \times 512$, generates approximately 255 tokens in GPT-4o. Metrics are evaluated using the Video Evaluation Agent. Tab.~\ref{tab_efficiency} highlights that adopting a 1 fps sampling ratio and a unified prompt reduces the cost of processing a 10-minute video from \$5.04 to \$0.83 while maintaining comparable performance to configurations with higher sampling rates and isolated prompts.

%-------------------------------------------------------------------------

\section{Fidelity Evaluation} 
For the quantitative experiments on video trimming, we also introduce a fidelity evaluation to assess the visual content similarity between the generated videos from different methods. For previous methods, we directly concatenate intervals with the highest saliency scores. In this experiment, we measure the feature similarity between the final video and the raw videos. A well-trimmed video should preserve the full content of the original video while maintaining its narrative coherence. 

We utilize two benchmarks, leveraging video features extracted by ViCLIP~\cite{wang2023internvid} and InternVideo2~\cite{wang2024internvideo2}. For both raw and trimmed videos from each method, an equal number of keyframes are sampled and processed through vision encoders. The feature similarity between the raw and trimmed videos is subsequently evaluated. As shown in Tab.~\ref{tab_viclip_similarity}, our method consistently improves content fidelity across various feature extraction models.

\begin{figure*}[t]
    \centering
    \includegraphics[width=\textwidth]{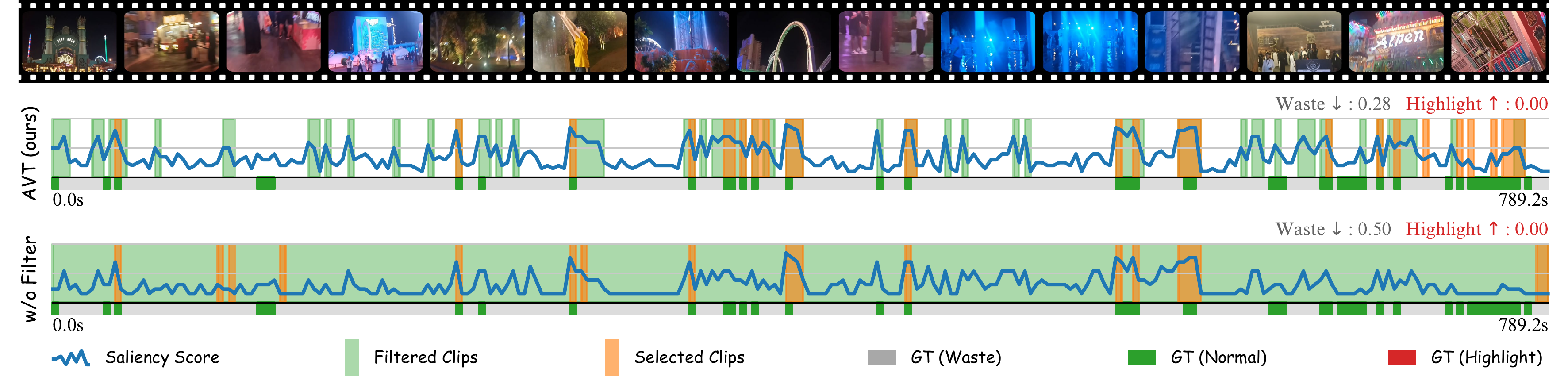}
    \vspace{0pt}
    \caption{
    Effect of clip filtering on visualization of trimmed videos.
    }
    \label{fig_vis_ablation_filter}
\end{figure*}

\begin{figure*}[t]
    \centering
    \includegraphics[width=\textwidth]{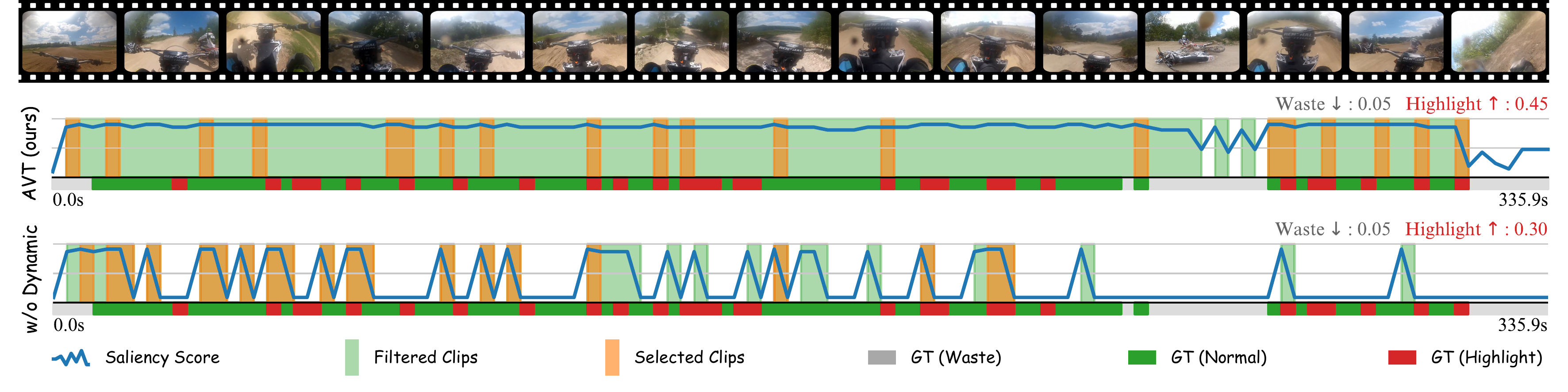}
    \vspace{0pt}
    \caption{
    Effect of dynamic filter module on visualization of trimmed videos.
    }
    \label{fig_vis_ablation_dynamic}
\end{figure*}%

\section{More Visualization}\label{more_vis}
In the main paper, we present visualizations of saliency scores and selected intervals for each method, demonstrating the effectiveness of our waste filtering operation and composition phase. In this supplementary section, we expand the analysis by incorporating additional visualizations and conducting case studies to highlight the significance of the AVT module designs.

\subsection{Clip selection}
We present a case study that visualizes the clip selection results when incorporating different AVT modules. Figure~\ref{fig_vis_ablation_filter} illustrates the impact of AVT's clip filtering module by comparing performance with and without it. Without filtering, story composition is applied to all intervals, resulting in a full row of light green segments in the visualization. This lack of candidate narrowing leads to the inclusion of more wasted footage in the final video.
Figure~\ref{fig_vis_ablation_dynamic} highlights the consequences of omitting the dynamic filtering module. Without this module, the clip filter discards most segments, especially in sports content, where intense activity often introduces jitter or other visual defects. As a result, highlight segments are misclassified as defects and excluded from the composition. The second row in the visualization shows significantly fewer filtered clips (light green) compared to the first row, emphasizing the importance of the dynamic filtering module.
The joint design of the AVT modules substantially enhances the viewing experience and enriches the content. By selecting more highlight footage and minimizing wasted footage, AVT not only outperforms prior approaches but also preserves a coherent storyline that aligns with the raw video material.

%------------------------------------------------------------------------------

\subsection{Storyline}
\begin{figure*}[ht]
    \centering
    \includegraphics[width=\textwidth]{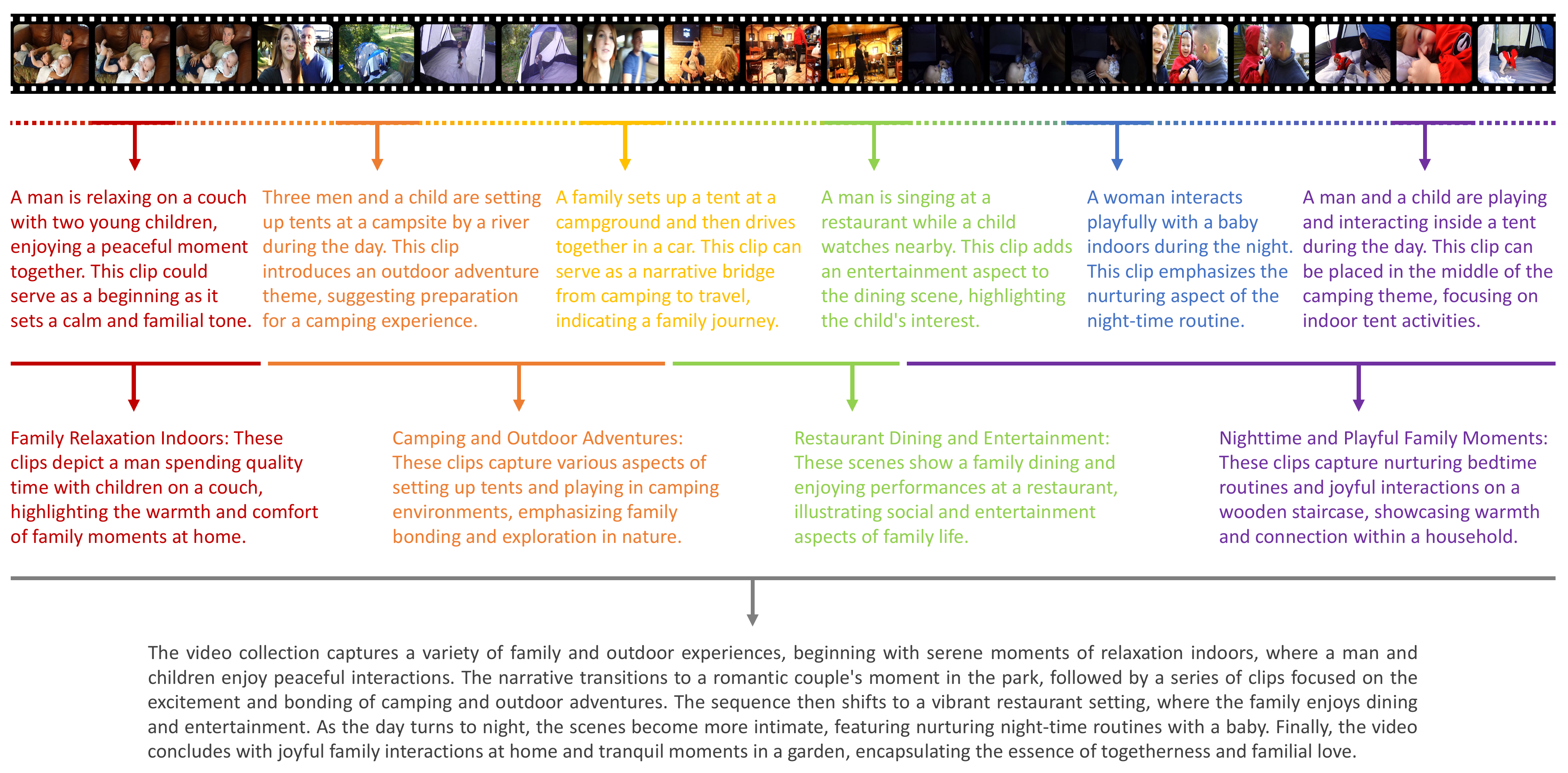}
    \vspace{0pt}
    \caption{
    Visualization of the multi-level storyline of the trimmed final video.
    }
    \label{fig_vis_captions}
\end{figure*}

We create the final video by constructing a corresponding storyline that outlines the rationale behind selecting each clip as the beginning, development, or ending, referred to as clip-wise captions. Additionally, we generate clustered themes, each representing a group of selected segments, as outlined in the story composition prompts. Ultimately, this results in a global storyline that captures the entire content of the trimmed videos. These captions, presented at various levels, are visualized in Fig.~\ref{fig_vis_captions}.

%------------------------------------------------------------------------------

\subsection{More Visualization with Existing Methods}
\begin{figure*}[ht]
    \vspace{-3pt}
    \centering
    \includegraphics[width=0.975\textwidth]{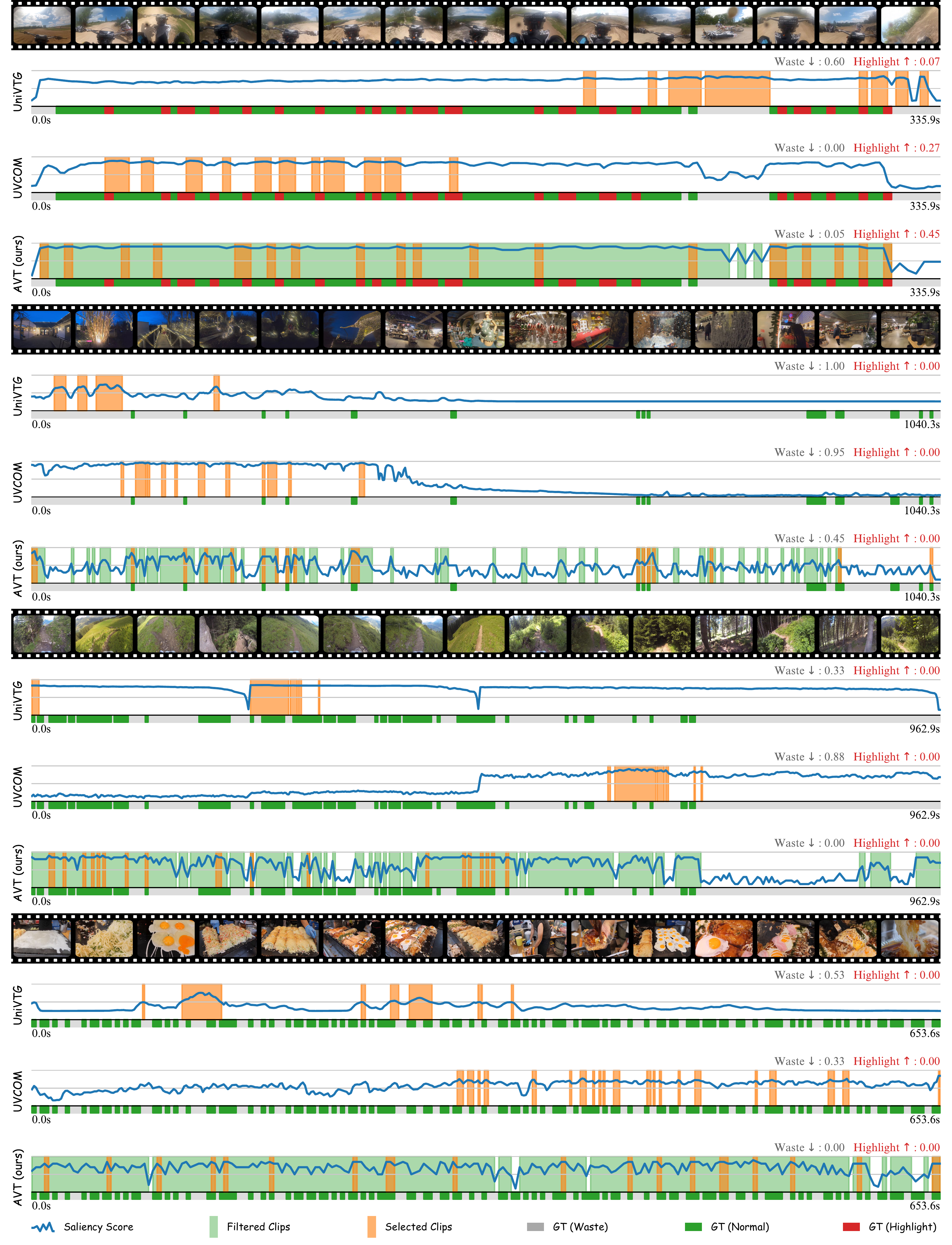}
    \vspace{-10pt}
    \caption{
    Visualization of trimmed videos on the video trimming dataset. 
    }
    \vspace{-5pt}
    \label{fig_more_vis}
\end{figure*}

We present additional visualizations of the saliency scores and selected intervals for each method in our video trimming dataset, as shown in Fig.~\ref{fig_more_vis}. Overall, AVT outperforms previous approaches by selecting more highlight footage, reducing wasted footage, and maintaining a consistent storyline, ultimately enhancing the viewing experience.

For instance, AVT excels at retrieving dynamic scenes like mountain biking, while existing methods tend to select more mundane clips. In another scenario, for plain vlogs such as food videos or dolphin shows, AVT efficiently trims the complete story across the entire timeline of the source video, while other methods may overlook key content.

\clearpage
\begin{figure*}[ht]
    \centering
    \includegraphics[width=0.975\textwidth]{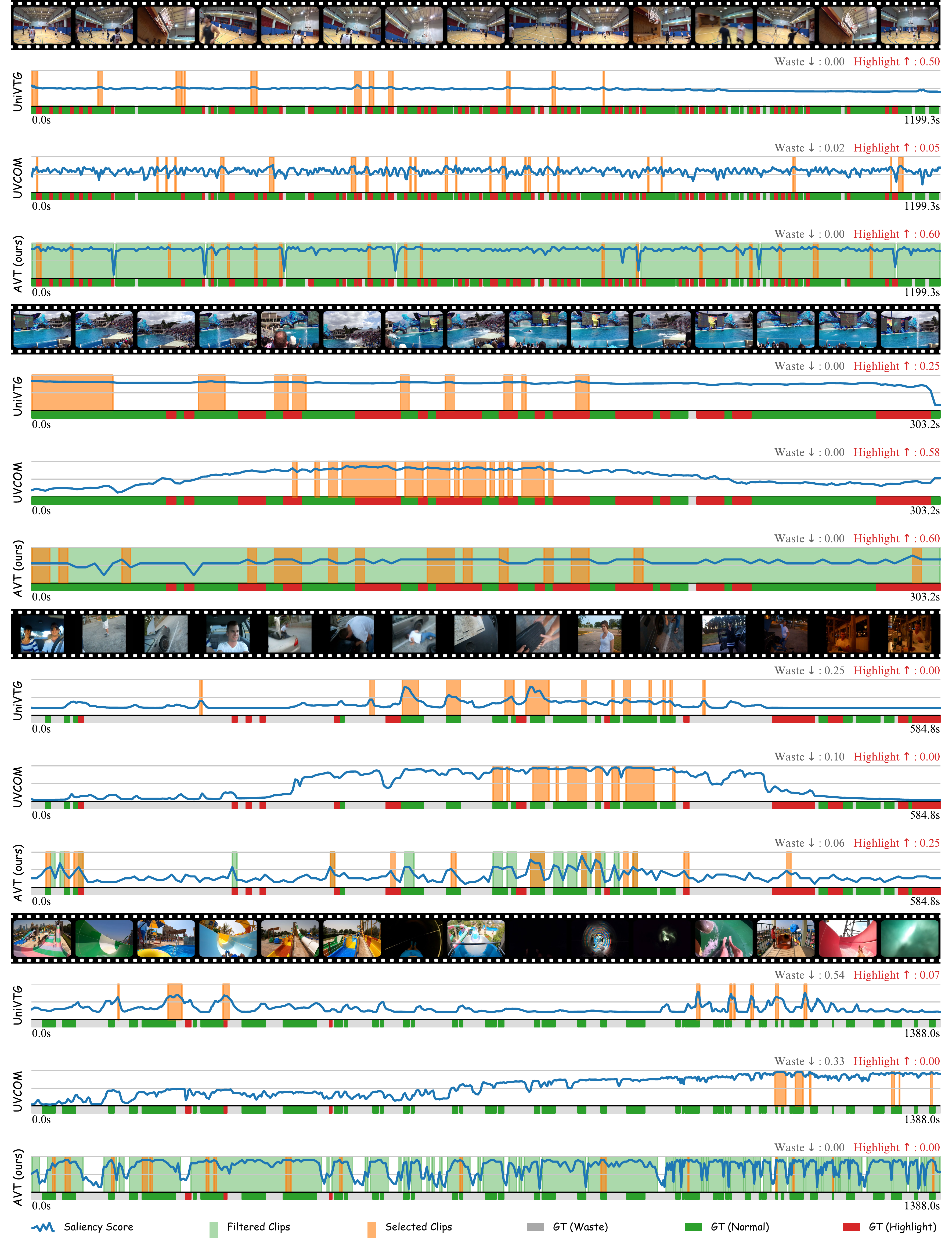}
    \label{fig_more_vis_1}
\end{figure*}
% \fi

\end{document}